\RequirePackage{xcolor}
\documentclass[pdflatex,sn-nature]{sn-jnl}

\usepackage{graphicx}
\usepackage{booktabs}
\usepackage{amsmath}
\usepackage{amssymb}
\usepackage{microtype}
\usepackage{xcolor}
\usepackage{hyperref}
\hypersetup{hidelinks}
\usepackage{fancyhdr}
\usepackage{eso-pic}
\usepackage{listings}

\AddToShipoutPictureBG*{%
  \AtPageUpperLeft{%
    \raisebox{-2.5cm}{\hspace*{\dimexpr0.5\paperwidth-0.5\textwidth\relax}%
      \colorbox{yellow!30}{%
        \parbox{\textwidth}{\centering\small\itshape
          Preprint --- not peer reviewed
        }%
      }%
    }%
  }%
}

\definecolor{promptbg}{gray}{0.96}
\definecolor{promptrule}{gray}{0.65}
\definecolor{placeholdercolor}{HTML}{B45309}   
\definecolor{commentcolor}{gray}{0.50}

\lstdefinestyle{promptjson}{
  basicstyle=\small\ttfamily,
  backgroundcolor=\color{promptbg},
  frame=single,
  rulecolor=\color{promptrule},
  breaklines=true,
  breakatwhitespace=false,
  columns=flexible,
  keepspaces=true,
  showstringspaces=false,
  tabsize=2,
  morecomment=[l]{//},
  commentstyle=\color{commentcolor}\itshape,
  moredelim=[s][\color{placeholdercolor}\bfseries]{<}{>},
  aboveskip=4pt,
  belowskip=4pt,
}

\graphicspath{{./}}

\title[LLM REALITY MONITORING]{Reality Monitoring in Large Language Models: Self-Knowledge
       That Transforms with Conversation Memory}

\author*[1]{\fnm{Saurabh} \sur{Ranjan}}\email{ranjan.saurabh@outlook.com}
\author[2]{\fnm{Konstantina} \sur{Sokratous}}
\author[1]{\fnm{Brian} \sur{Odegaard}}

\affil*[1]{\orgdiv{Department of Psychology}, \orgname{University of Florida},
           \orgaddress{\city{Gainesville}, \state{Florida}, \country{USA}}}
\affil[2]{\orgname{University of Missouri},
          \orgaddress{\state{Missouri}, \country{USA}}}

\abstract{%
A conversational AI that cannot tell its own output from what a user
said will treat its own mistakes as user-provided facts.  In humans,
this capacity is called reality monitoring, and its failures are
linked to hallucinations, delusions, and confabulation, yet whether
LLMs possess it remains untested.  Here we show, across two
experiments and six LLMs, that source attribution depends on how
conversational memory is structured: ceiling accuracy for
self-generated content under minimal memory demands reverses to a
fragile external-item advantage once episodic delay removes that
shortcut.  Feedback exposes two failures: in some models, internal
and external judgments swap; in others, accuracy improves while
confidence decouples from correctness, dissociations invisible to
existing benchmarks.  Across models, this pattern implicates active,
not aggregate, parameter count.  This suggests that as AI systems
take on autonomous, multi-turn roles, evaluating what they know is
not enough: tracking where that knowledge came from may matter
equally.%
}


\begin{document}
\maketitle



Large language models (LLMs) now achieve human-competitive performance
across a broad range of language generation and reasoning tasks
\cite{brown2020language,wei2022emergent,chang2024llms,petroni2019language,%
hernandez2024linearity,li2024emergent},
prompting investigation of whether higher-order cognitive abilities
such as introspection, self-modeling, and situational awareness emerge
in these systems~\cite{lindsey2025emergent,binder2024looking,long2023introspective,%
comsa2025introspection,chen2025imitation,laine2023situational,nguyen2025probing,%
needham2025llms,apollo2025claude,anthropic2024claude3}.

\begin{figure}[htbp]
  \centering
  \includegraphics[width=\linewidth]{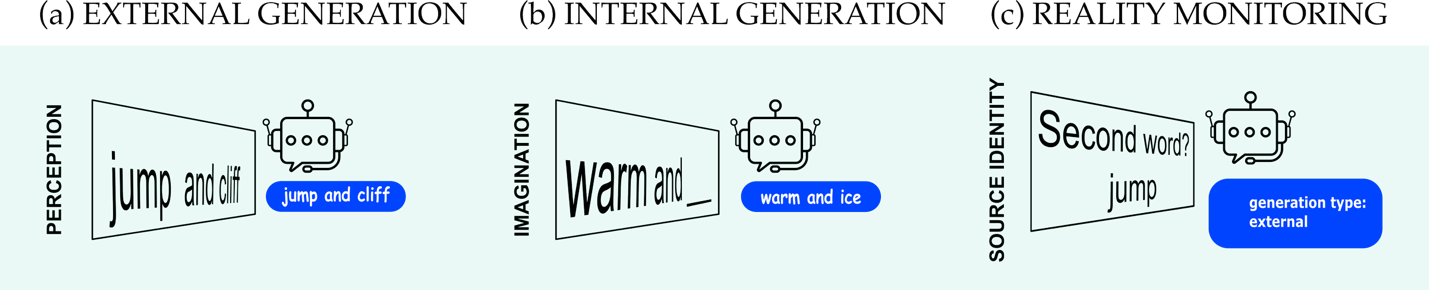}
  \caption{\textbf{A word-pair paradigm operationalizes reality
           monitoring in LLMs by contrasting externally supplied and
           self-generated second words.}
           (a) \textit{External generation:} the LLM receives a complete
           word pair (first word and second word) and must reproduce the
           second word verbatim.  This simulates perception of an external stimulus: the word exists outside the model and requires
           accurate reproduction rather than generation.
           (b) \textit{Internal generation:} the LLM receives only the
           first word of a pair and must produce a novel second word.
           The generated word exists only within the model's own output, a simulation of internally generated content.  No constraint is
           placed on the choice of word beyond it being a single English
           word not previously used in the session.
           (c) \textit{Reality monitoring:} the model must determine whether the
           associated information was externally provided (external/perceived) or
           self-generated (internal/imagined).}
  \label{fig:concept}
\end{figure}

Whether LLMs can discriminate information that is \textit{externally
generated} (presented by a user) from information that is
\textit{internally generated} (produced by the model itself) has not
been systematically examined using paradigms permitting direct
human--LLM comparison or isolating conversational memory as an
experimental variable (Figure~\ref{fig:concept}).  Yet LLMs
increasingly operate in high-stakes agentic contexts, including
clinical decision support \cite{singhal2023large}, legal analysis
\cite{choi2023chatgpt}, autonomous scientific reasoning
\cite{boiko2023autonomous}, and military decision-making
\cite{rivera2024escalation}, where they both generate and receive
information across extended multi-turn interactions.  A model that
cannot attribute information to its correct source may cite
confabulations as user-provided facts, fail to update when corrected,
or misreport its own claims' epistemic authority: source-attribution
failure is not a performance error but a safety failure.

In cognitive science, \textit{reality monitoring} (RM;
Figure~\ref{fig:concept}) is the capacity to determine the origin of a
memory: whether its content derives from something externally perceived
or something internally imagined \cite{johnson1981reality,johnson1988reality}.
Human RM is imperfect but systematic: internally generated information
tends to carry richer cognitive operations and fewer perceptual details,
whereas externally generated information has more perceptual specificity
\cite{johnson1981reality,johnson1988reality,garrison2017monitoring,%
simons2017brain,simons2022brain,simons2008separable,dijkstra2022perceptual,%
aleman2004fantasy,aynsworth2017reality,bentall1991reality},
and on word-pair materials humans show systematic externalizing biases
(i.e., often confusing imagined information as previously perceived)
and source-dependent metacognitive
sensitivity~\cite{ranjan2024reality}.
Failures of RM have been associated with hallucinations and delusions in
schizophrenia
\cite{alpert1985signs,aleman2003cognitive,brebion1997discrimination,%
brebion2002source,brunelin2006source,brunelin2007impaired,thoresen2014frontotemporal},
delusional disorders
\cite{thoresen2014frontotemporal,corlett2009illusions,venneri2000nurturing},
paranoia
\cite{benedetti2005reality,koller2021paranoia,rossigoldthorpe2021paranoia,%
corlett2019hallucinations},
and confabulation
\cite{corlett2014dreams,dallabarba2017confabulations,schnider2001spontaneous,%
fotopoulou2007confabulation}.

Despite its relevance, RM has not been systematically investigated in
LLMs, although LLM hallucinations
\cite{alansari2025hallucination,huang2025survey} (outputs that do not correspond to verifiable external states) are conceptually
adjacent to RM failures.  In particular, \textit{conversational
memory}, which determines which sources are even available for
attribution at test, has never been isolated as an experimental
variable in source attribution research (see Related Work).

\begin{figure}[htbp]
  \centering
  \includegraphics[width=\linewidth]{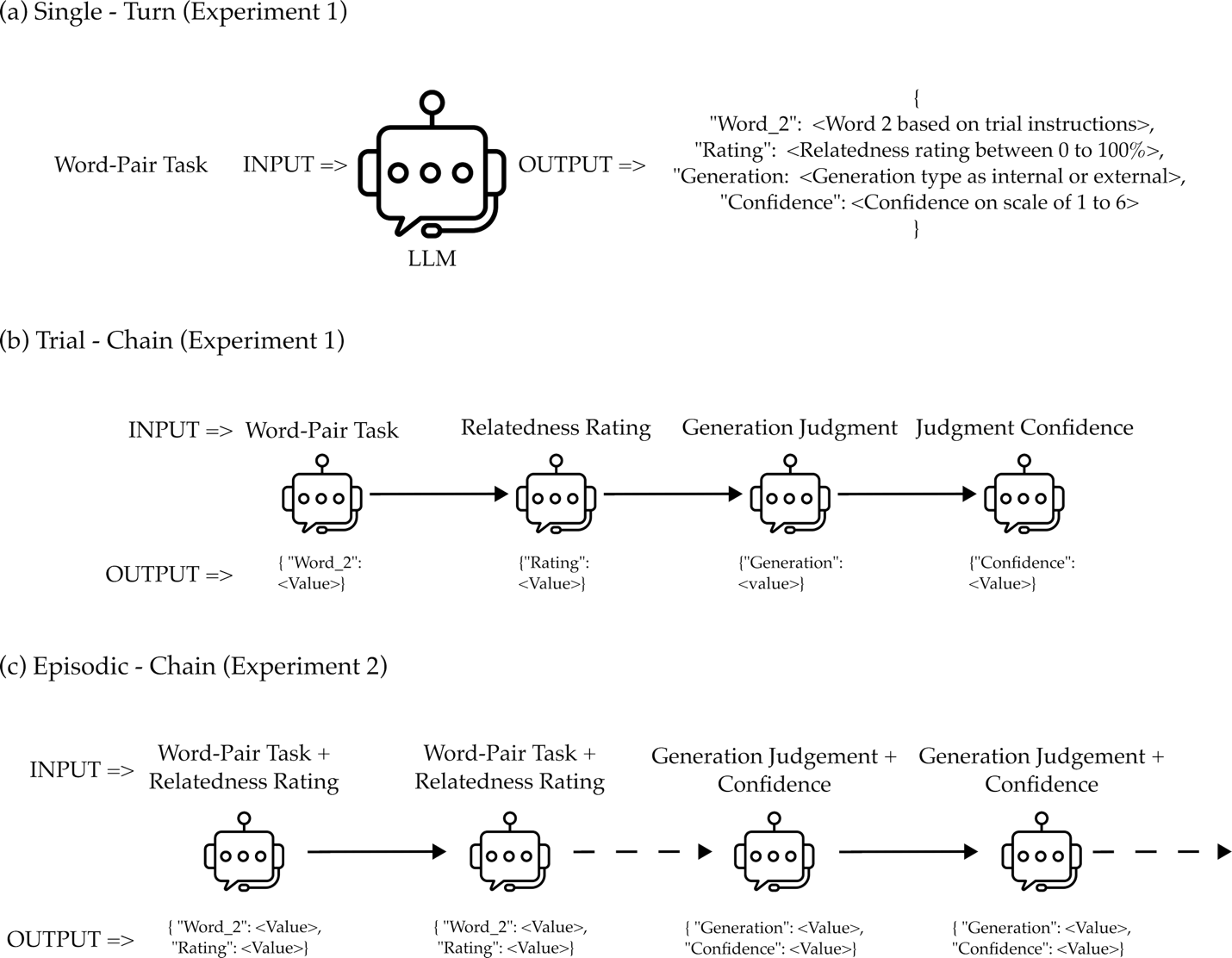}
  \caption{\textbf{Three conversational memory architectures vary how much
           prior context a model must maintain across turns to complete
           source attribution.}
           (a) \textit{Single-Turn:} all four subtasks --- word-pair
           completion, semantic relatedness rating (0--100 scale), source
           judgment (internal or external), and confidence rating (1--6)
           --- are requested within a single LLM interaction.  The model
           encodes and attributes the source of each trial without carrying
           information across separate turns.
           (b) \textit{Trial-Chain:} the same four subtasks are distributed
           across four sequential conversational turns, each extending the
           running conversation history.  Source attribution occurs in the
           isolated interaction with the LLM, after the completion and relatedness rating have
           already been provided in past interaction.  This tests whether decomposing
           within-turn cognitive load improves source classification
           accuracy, while keeping the total number of trials identical
           to Single-Turn. The Single-Turn and Trial-Chain conditions are examined in Experiment 1.
           (c) \textit{Episodic-Chain:} all encoding trials (5 or 10
           word pairs per source type, yielding set sizes of 20 or 40
           total trials including test trials) are completed as a single block before a
           separate delayed source-test phase begins.  Source attribution
           is probed only after the full encoding sequence, introducing
           genuine episodic delay analogous to the between-phase interval
           in human RM experiments.  Corrective feedback (present
           vs.\ absent after each source response) is crossed with set
           size to yield four conditions (a $2 \times 2$ feedback-by-set-size
design; source type, internal vs.\ external, is analyzed as a separate
within-trial factor, not part of this crossing). Episodic-Chain with set size and feedback manipulations is examined in Experiment 2.}
  \label{fig:impl}
\end{figure}

Here, we adapt the RM framework from human cognitive psychology to
assess source monitoring in LLMs.  In a word-pair task, LLMs encode
pairs whose second word is either externally provided (perceived) or
internally generated (imagined), and then are asked to attribute the
source of the associated second word as internal/imagined or
external/perceived, enabling direct comparison with human
RM~\cite{ranjan2024reality,johnson1981reality,johnson1988reality}.
Across two experiments, we assess conversational memory as our
primary experimental variable of interest (Figure~\ref{fig:impl}).
Experiment~1 varies within-trial
memory structure: Single-Turn, in which all subtasks occur within one
conversational turn, vs.\ Trial-Chain, in which the same subtasks are
distributed across sequential turns.  Experiment~2 introduces an
Episodic-Chain paradigm mirroring human episodic RM tasks: encoding
trials are completed as a single block before a separate, delayed
source-test phase, with set size (20 vs.\ 40 trials) and corrective
feedback (present vs.\ absent) crossed to test whether AI
self-knowledge, the capacity to know what the model itself generated,
is stable across memory loads and whether feedback can recalibrate
it.  We evaluate six LLMs spanning Gemma~3
(12B and 27B, with and without quantization-aware
training), Llama~3.3 (70B), and a sparse mixture-of-experts model
(Llama~4 16$\times$17B), using the same paradigm and materials as a
validated human RM dataset.

\section{Related Work}

Human RM research has established that source attribution accuracy
and metacognitive calibration are systematically modulated by stimulus
relatedness, source type, and episodic
delay~\cite{johnson1981reality,johnson1988reality,ranjan2024reality};
the human behavioral profile for a similar word-pair stimulus
set~\cite{ranjan2024reality} provides a qualitative baseline here.
Source identification in AI systems has largely been addressed through
external mechanisms such as
watermarking~\cite{huang2025survey,alansari2025hallucination} rather
than probing intrinsic self-attribution capacity.  Theoretical work
links higher-order representation, a system's capacity to represent
its own inner representational states, to metacognition and
self-knowledge, and argues that whether LLMs form such representations
remains empirically unresolved~\cite{butlin2026higher}; empirical work
on LLM introspection, situational awareness, and self-modeling raises
the same
concern~\cite{laine2023situational,long2023introspective,comsa2025introspection,nguyen2025probing},
but none has used formal, behaviorally grounded RM paradigms or
treated conversational memory as a manipulated variable.  The most
closely related benchmarks test metacognitive calibration on knowledge
and reasoning tasks~\cite{wang2026mirror,cacioli2026metacognitive},
where frontier LLMs show above-chance but limited-resolution,
context-dependent ability~\cite{ackerman2026limited}; work evaluating
episodic memory with sequence-order recall
tasks~\cite{pink2024assessing} has likewise not examined source
attribution or metacognitive sensitivity.  None of this work uses
source-memory paradigms permitting direct human--LLM comparison, nor
treats conversational memory architecture as a controlled variable.

\section{Results}

\begin{figure}[htbp]
  \centering
  \includegraphics[width=\linewidth]{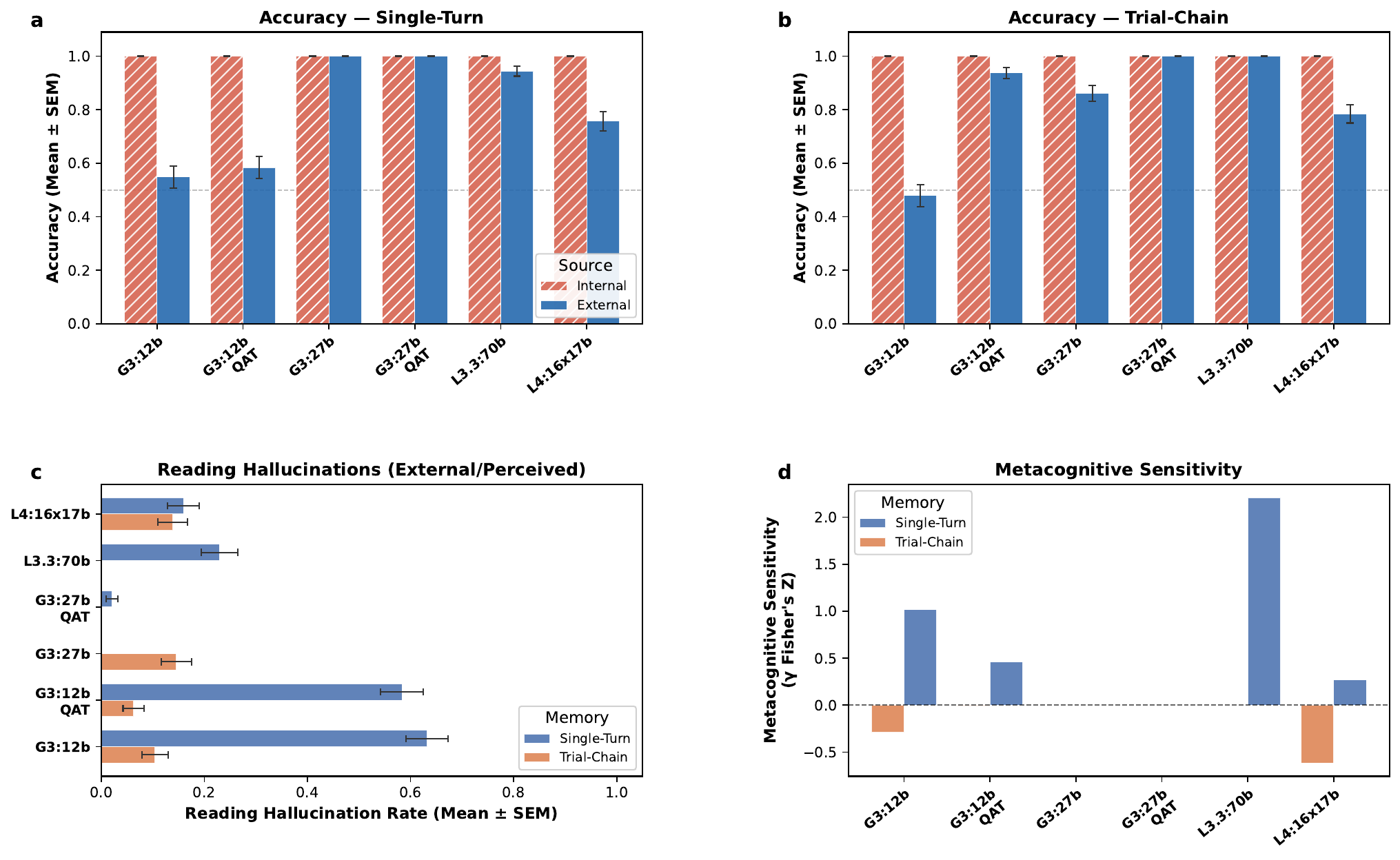}
  \caption{\textbf{Internal source identification reaches ceiling across all
           models; external accuracy varies with reading hallucination
           rates and model scale.}
           Performance metrics for single-trial reality monitoring across
           six LLMs (Gemma3:12b, Gemma3:12b-QAT, Gemma3:27b,
           Gemma3:27b-QAT, Llama3.3:70b, Llama4:16x17b) and two memory
           implementations in Experiment~1.
           (a--b) Mean source attribution accuracy for the Single-Turn~(a)
           and Trial-Chain~(b) conditions, shown separately for internal
           (red, hatched) and external (blue, solid) trials; error bars
           $\pm$1~SEM across trials within each model--condition cell.
           (c) Mean reading hallucination rates on external trials only
           ($N = 1{,}728$ external trials; missing bars indicate zero
           hallucinations observed in that cell).
           (d) Metacognitive sensitivity (Goodman--Kruskal $\gamma$,
           Fisher $Z$-transformed) relating expressed confidence to
           trial-level accuracy, computed once per model--condition cell
           from all trials in that cell pooled (not an average of
           per-trial estimates; no error bars defined).  Positive
           $\gamma$ indicates above-chance calibration.  Missing bars
           indicate $\gamma$ was undefined because ceiling accuracy left
           zero variance in the correctness variable.}
  \label{fig:exp1}
\end{figure}

Experiment~1 yielded ceiling performance for internal items but
variable external-item accuracy; Experiment~2 introduced episodic
delay and reversed the source advantage
(Figures~\ref{fig:exp1}--\ref{fig:sdt}).

\subsection{Experiment 1: Single-Trial Reality Monitoring}

\subsubsection{Accuracy: Ceiling-Level Performance for Internal Items, Variable Performance for External Items}

Across all models and both implementations (Figure~\ref{fig:exp1}),
internally generated (imagined) items were identified with perfect
accuracy (100\%).  This ceiling reflected direct within-turn
retrieval: the generated word remained in the active context window at
the time of judgment, making imagined-item identification trivially
accessible rather than requiring inferential source monitoring.
Externally generated (perceived) items, by contrast, showed
substantial variability across models and implementations.

A generalized linear model over perceived trials ($N = 1{,}728$;
predictors: model, implementation, response-option order, relatedness,
reading hallucination; pseudo $R^2 = .474$; Methods) identified reading
hallucination as the strongest predictor of source attribution failure
($b = -3.84$, $p < .001$): a model that failed to reproduce the
externally presented word was substantially more likely to misclassify
its source, establishing reading hallucinations as a primary
bottleneck for RM on externally generated trials.  Conditional on no
reading hallucination, predicted accuracy was near ceiling (90.9\% for
Gemma3:12b, the lowest-scoring model; 95.2\%--99.7\% for the remaining
reliably estimated models), indicating that source monitoring itself
remains largely intact once this upstream encoding failure is excluded
(Gemma3:27b-QAT's coefficient is unreliable: quasi-complete
separation; Methods).

Model scale also mattered: larger models generally achieved higher
accuracy (Llama3.3:70b: 94.4\% Single-Turn, 100\% Trial-Chain), while
smaller models could fall to chance (Gemma3:12b: 47.9\% Trial-Chain).
Trial-Chain outperformed Single-Turn for three of six models, tied at
ceiling for one, and reversed for two (Gemma3:12b, Gemma3:27b).  For
Gemma3:27b this reflected a hallucination rate that rose under
Trial-Chain (0\% to 14.6\%); for Gemma3:12b, hallucination rates fell
sharply (63.2\% to 10.4\%) yet accuracy still declined, confirming a
genuine shift in source-judgment ability independent of encoding
fidelity (conditional on correct word reproduction, accuracy fell from
100\%, $n = 53$, in Single-Turn to 53.5\%, $n = 129$, in Trial-Chain,
barely above chance).  Reduced cognitive load improves RM for some
architectures but not others.

\subsubsection{Reading Hallucinations}

A GLM with the same predictors ($N = 1{,}728$ perceived trials)
showed that Trial-Chain significantly reduced hallucination rates
relative to Single-Turn ($b = -1.80$, $p < .001$): decomposing the
task across conversational turns improves fidelity to externally
supplied inputs.  More related word pairs produced slightly fewer
hallucinations ($b = -0.006$, $p = .036$), and response-option order
reached significance (odds ratio [OR] $= 1.40$, $p = .021$).  All four larger
models hallucinated significantly less than the Gemma3:12b baseline
(Gemma3:27b-QAT: $b = -4.30$; Gemma3:27b: $b = -2.27$; Llama3.3:70b:
$b = -1.75$; Llama4:16x17b: $b = -1.49$; all $p < .001$), whereas the
12B quantization-aware variant did not differ ($b = -0.22$,
$p = .24$).

\subsubsection{Metacognitive Sensitivity: Single-Turn Advantage}

Metacognitive sensitivity, the extent to which an agent's confidence predicts its accuracy \cite{fleming2014measure}, was
estimated per model and implementation using the Goodman--Kruskal
$\gamma$ coefficient (Figure~\ref{fig:exp1}d).  Single-Turn generally
yielded stronger positive confidence--accuracy associations than
Trial-Chain: Gemma3:12b, for example, showed $\gamma = .77$ in
Single-Turn but $\gamma = -.28$ in Trial-Chain, and several other
models showed similar or uncomputable reversals (the latter when
ceiling or floor accuracy left no variance in correctness).  Reducing
cognitive load for task performance therefore does not necessarily
preserve confidence calibration.

\subsubsection{Confidence in Single-Trial Reality Monitoring}

Confidence ratings in Experiment~1 were modeled using ordered probit
regression.  Perceived-source trials elicited substantially higher
confidence than imagined-source trials ($b = 1.80$, $p < .001$),
reading hallucinations reduced confidence ($b = -0.67$, $p < .001$), indicating greater uncertainty when reproduction of the external word failed, and trial-level accuracy independently
increased confidence ($b = 1.05$, $p < .001$).  Trial-Chain was
associated with higher confidence than Single-Turn ($b = 1.28$,
$p < .001$), likely reflecting the more structured context window.

\subsection{Experiment 2: Episodic Reality Monitoring}

Experiment~2 introduced delayed source testing across an extended
conversational history, crossed with set size (20 vs.\ 40 trials)
and feedback (absent vs.\ present).

\begin{figure}[htbp]
  \centering
  \includegraphics[width=\linewidth]{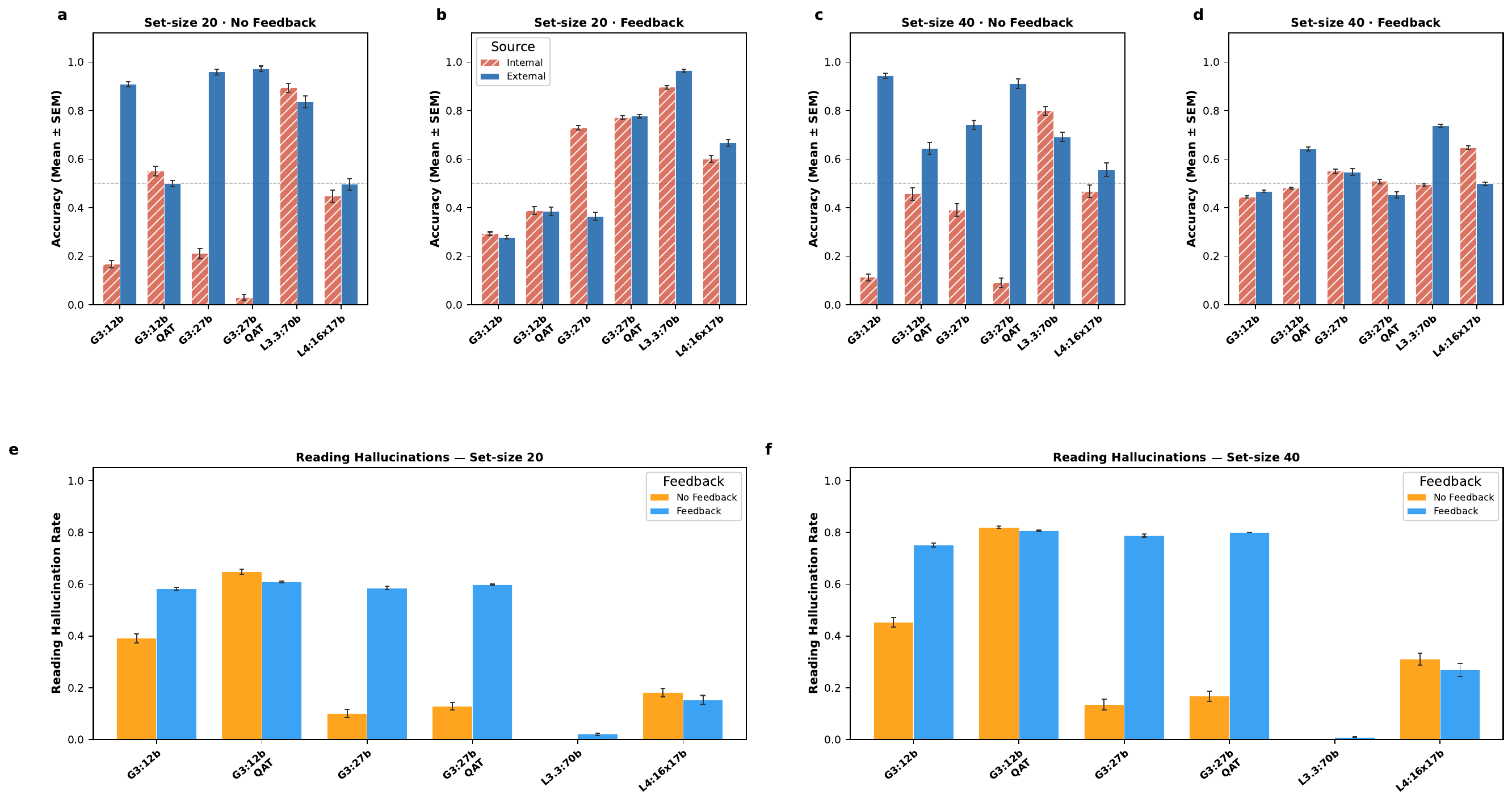}
  \caption{\textbf{Episodic delay reverses the source advantage for
           internally-generated words, and feedback selectively
           recalibrates internal-item accuracy at the cost of increased
           reading hallucinations.}
           Performance in the Episodic-Chain paradigm (Experiment~2).
           (a--d) Mean source attribution accuracy for internal (red,
           hatched) and external (blue, solid) trials across four
           conditions: (a)~set size~20, no feedback; (b)~set size~20,
           feedback; (c)~set size~40, no feedback; (d)~set size~40,
           feedback.  Error bars: $\pm$1~SEM across trace IDs, each
           trace treated as an independent participant ($N = 200$ per
           cell); Gemma3:12b's response inversion under feedback is
           detailed in Figure~\ref{fig:sdt}.
           (e--f) Mean reading hallucination rates on external trials
           for set size~20~(e) and set size~40~(f); bars show
           no-feedback (orange) vs.\ feedback (dark blue).}
  \label{fig:exp2}
\end{figure}

\subsubsection{Accuracy: Reduced Performance and Source Reversal}

In Experiment~2, no model reached ceiling (Figure~\ref{fig:exp2}).
Mean accuracy (trace-level, $N = 4{,}800$ traces per source) was
higher for perceived items ($M = 66.42\%$,
range across conditions: 27.80\%--97.20\%) than for imagined items
($M = 47.57\%$, range: 3.00\%--89.60\%), a source reversal relative to
Experiment~1, where imagined items had been easier to attribute.  The
reversal was most pronounced without feedback (perceived $M = 76.34\%$
vs.\ imagined $M = 38.46\%$); feedback nearly eliminated the gap
($M = 56.50\%$ vs.\ $M = 56.68\%$), with heterogeneous effects on
individual models, including systematic response inversion in
Gemma3:12b (Signal Detection Analysis, below).

A generalized linear mixed model (GLMM; binomial family, logit link;
random intercept for trace; $N = 72{,}000$ test-phase observations;
Methods) confirmed the reversal: perceived items were attributed more
accurately than imagined items ($\chi^2(1) = 2594.9$, $p < .001$;
OR $= 2.57$, 95\% CI $[2.48,\, 2.66]$).  The source $\times$ feedback
interaction was pronounced ($\chi^2(1) = 2734.1$, $p < .001$):
feedback differentially improved imagined-item accuracy while
simultaneously increasing reading hallucinations on perceived trials,
attenuating the source advantage.  Unlike Experiment~1, restricting to
hallucination-free perceived trials did not restore near-ceiling
accuracy (no feedback: $79.1\%$ vs.\ $67.9\%$ with hallucination;
feedback: $53.4\%$ vs.\ $58.8\%$): episodic-delay failures are not
reducible to encoding errors, so source monitoring itself degrades
under memory load.  Main effects of set size and
feedback and their interaction were also significant (all
$p < .001$), as was the source $\times$ set size interaction
($p = .010$), though the latter was negligible in practical terms
(OR $= 0.91$).

\subsubsection{Reading Hallucinations}

A GLMM predicting reading hallucinations ($N = 36{,}000$ perceived
trials; binomial family, logit link) showed that hallucinations
increased with larger set sizes (OR $= 1.56$, $p < .001$) and feedback
exposure (OR $= 3.72$, $p < .001$), amplified in larger sets under
feedback (Type-II Wald $\chi^2(1) = 19.8$, $p < .001$).  Model
differences were pronounced: relative to baseline, Llama3.3:70b
(OR $= 0.002$) and Llama4:16x17b (OR $= 0.15$, both $p < .001$) showed
dramatically suppressed hallucination rates, whereas Gemma3:12b-QAT
hallucinated at higher odds than its non-QAT counterpart
(OR $= 2.64$, $p < .001$), reversing the Experiment~1 pattern, in
which no quantization-aware variant exceeded baseline.  All other
coefficients are in the Supplementary Information.

\subsubsection{Metacognitive Sensitivity: Degradation Under Memory Load and Feedback}

\begin{figure}[htbp]
  \centering
  \includegraphics[width=\linewidth]{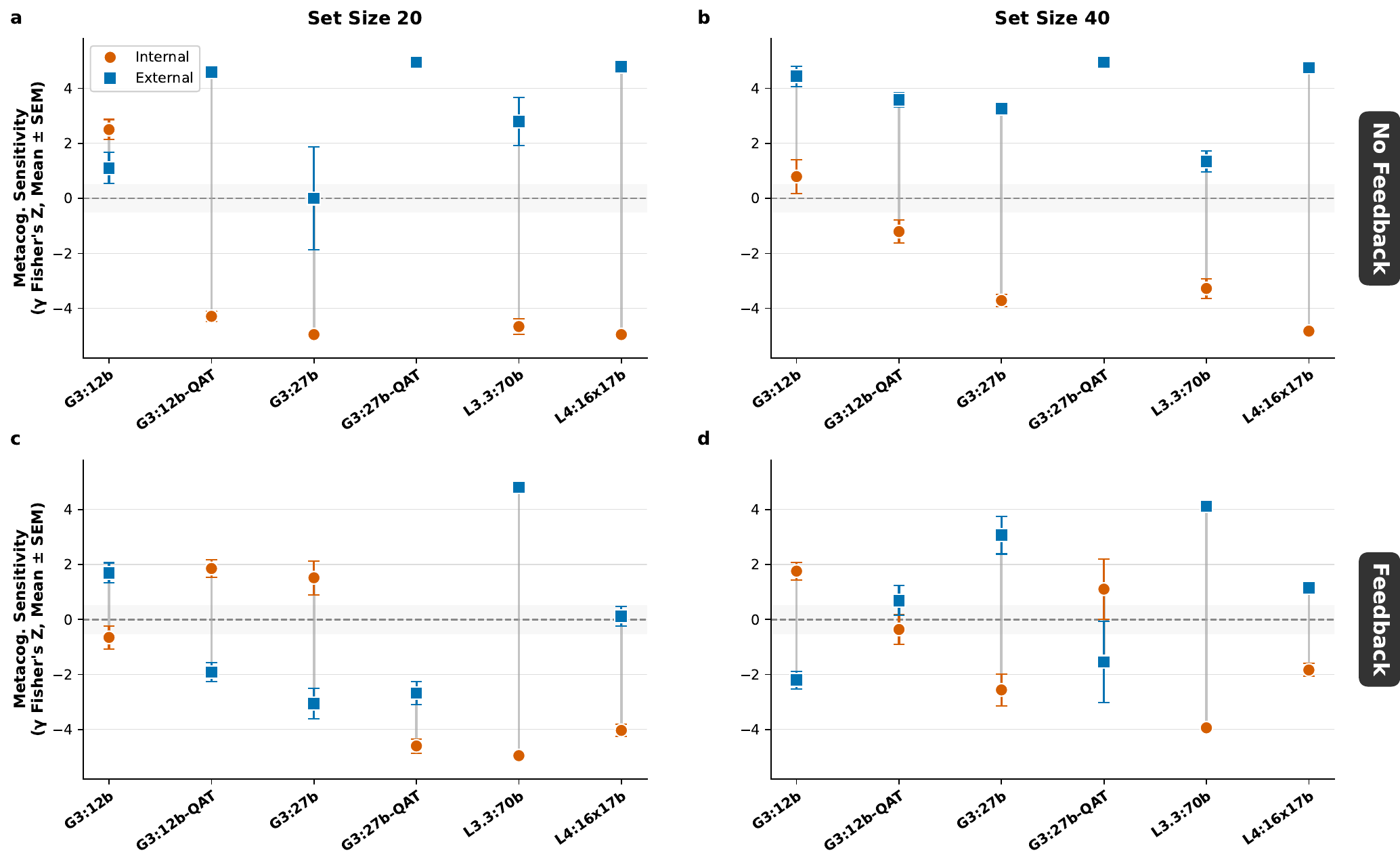}
  \caption{\textbf{Metacognitive sensitivity is consistently higher for
           external than internal items and is substantially reduced
           by corrective feedback and larger memory loads.}
           Fisher $Z$-transformed Goodman--Kruskal $\gamma$, a correlation between expressed confidence and trial-level accuracy, measuring metacognitive sensitivity separately
           for internal and external items across Episodic-Chain
           conditions in Experiment~2.  Positive $\gamma$ indicates
           above-chance calibration; $\gamma = 0$ indicates expressed
           confidence is unrelated to whether the source judgment was
           correct.
           Panels are arranged in a 2~(feedback) $\times$ 2~(set size)
           grid: top row~= no feedback; bottom row~= feedback; left
           column~= set size~20; right column~= set size~40.  Within
           each panel, orange circles~= internal items; blue
           squares~= external items; vertical grey lines
           connect the two source estimates for each model, making the
           source dissociation visible across conditions.  Error bars:
           $\pm$1~SEM across trace IDs, each trace treated as one
           participant.
           Corrective feedback markedly reduced external-item $\gamma$
           in most models, with the sharpest collapses in the Gemma
           family (largest: Gemma3:27b-QAT); Llama3.3:70b is the
           exception, with external-item $\gamma$ increasing under
           feedback even as its all-trials Type~2 AUC falls below
           chance (Figure~\ref{fig:sdt}), the latter reflecting
           cross-source confidence differences rather than
           within-source calibration.  Gemma3:27b-QAT internal-source
           $\gamma$ is absent in no-feedback panels (dots omitted); see
           Results for explanation.}
  \label{fig:metacog}
\end{figure}

Metacognitive sensitivity in episodic RM (Figure~\ref{fig:metacog}) was
analyzed using a linear mixed model (LMM) with random intercepts for
trace ($N = 4{,}410$ simulations with computable $\gamma$ values;
approximately 5{,}190 of 9{,}600 possible observations were excluded
due to ceiling or floor accuracy or invariant confidence; exclusion
pattern detailed in Methods).

Perceived items showed substantially higher metacognitive sensitivity
than imagined items (pairwise contrast perceived $-$ imagined
$= 4.28$, $SE = 0.12$, $p < .001$), and feedback \textit{decreased}
sensitivity ($b = -3.22$, $F(1, \infty) = 30.4$, $p < .001$),
amplified at the larger set size (set size $\times$ feedback:
$F(1, \infty) = 7.80$, $p = .005$), consistent with feedback and
extended context jointly eroding confidence--accuracy calibration;
set size otherwise improved calibration only for perceived items
(source $\times$ set size: $F(1, \infty) = 6.41$, $p = .011$).
Metacognitive performance was highly model-dependent, with
several models showing substantially lower $\gamma$ under feedback.
One model, Gemma3:27b-QAT, contributed no computable imagined-source
$\gamma$ without feedback: every trace's internal-source accuracy was
invariant, uniformly incorrect in 376 of 400 traces (a strong
external-response bias) and uniformly correct in the remainder,
leaving no within-trace variance to compute $\gamma$ (missing points
in Figure~\ref{fig:metacog}).  To identify what drives confidence when
accuracy does not, we modeled raw confidence directly.

\subsubsection{Confidence: Driven by Architecture and Context, Not Trial-Level Accuracy}

Ordinal confidence ratings (1--6) were analyzed using a Cumulative
Link Model \cite{christensen2023ordinal} (CLM; logit link;
$N = 72{,}000$; Methods).  Model identity was by far the dominant
predictor of confidence ($\chi^2(5) = 17{,}620.6$, $p < .001$), with
odds ratios relative to the Gemma3:12b baseline ranging from 0.44
(Llama4:16x17b) to 17.85 (Gemma3:27b-QAT), a 40-fold spread from
architecture alone.  Imagined items received slightly higher
confidence than perceived items (OR $= 1.10$, $p = .003$); feedback
(OR $= 0.86$, $p < .001$) and larger set size (OR $= 0.94$, $p = .048$)
reduced confidence, and higher relatedness slightly increased it
(OR $= 1.01$/unit, $p < .001$).  Three interactions qualified these
effects: feedback disproportionately reduced confidence for imagined
items (source $\times$ feedback: OR $= 0.60$, $p < .001$) and at set
size 40 (setsize $\times$ feedback: OR $= 0.56$, $p < .001$), whereas
the imagined-item advantage \textit{widened} at set size 40
(source $\times$ setsize: OR $= 1.15$, $p < .001$).  Full
coefficients, model comparisons, and post-hoc contrasts are in the
Supplementary Information.

Critically, trial-level accuracy did not predict confidence after
controlling for all other factors ($b = 0.03$, $p = .063$), nor did
reading hallucinations ($p = .985$), unlike Experiment~1, where
accuracy significantly predicted confidence.  A trace-level robustness
test confirmed this null (paired within-trace difference $= +0.001$
confidence points, $t(4518) = 0.11$, $p = .91$; Methods) and revealed
that it conceals two small opposite-signed effects: confidence weakly
tracked correctness without feedback ($+0.06$) but reversed under
feedback ($-0.06$; both $p < .001$).  Episodic delay thus specifically
erodes the accuracy--confidence link preserved under single-trial
memory, and feedback inverts it, consistent with the near- and
below-chance Type~2 AUC values below (Figure~\ref{fig:sdt}): expressed
confidence here reflects model architecture and condition, not
whether the individual source judgment was correct.

\subsubsection{Cumulative Accuracy: Heterogeneous Trial-Level Trajectories}

Cumulative accuracy trajectories across episodic test trials
(Supplementary Figure~S6) revealed substantial trial-level
heterogeneity not captured by mean accuracy alone: trajectory shape
depended jointly on model identity and set size rather than model
scale, with no single shape dominating either feedback condition,
indicating model-specific constraints on memory and context management
rather than a static capability limit.

\subsubsection{Signal Detection Analysis: Source Discrimination vs.\
               Metacognitive Calibration Dissociation}

\begin{figure}[htbp]
  \centering
  \includegraphics[width=\linewidth]{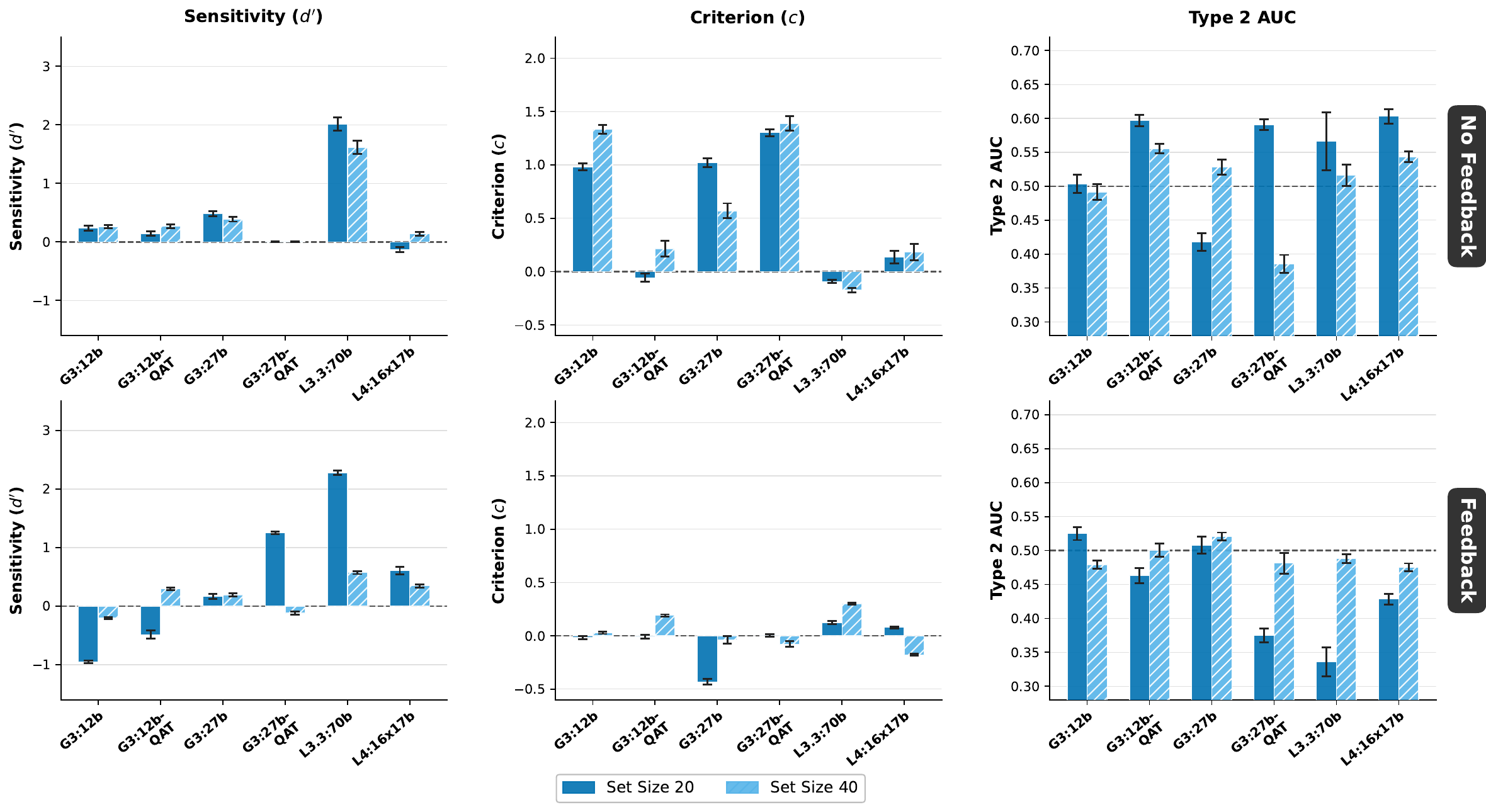}
  \caption{\textbf{Corrective feedback dissociates source discrimination
           from metacognitive calibration and produces qualitatively
           distinct failure modes by model scale.}
           Type~1 (source discrimination) and Type~2 (metacognitive)
           signal detection measures across the four Episodic-Chain
           conditions.  $d'$~= standardized source-discrimination
           sensitivity; criterion $c$~= response bias (negative values
           indicate a liberal bias toward the internally generated
           response); Type~2 AUC~= area under the receiver operating
           characteristic curve discriminating correct from incorrect
           source judgments, a model-free measure of metacognitive
           accuracy that does not depend on a single criterion placement.
           Two rows $\times$ three columns: rows~= feedback condition
           (top~= no feedback; bottom~= feedback); columns~= $d'$,
           criterion $c$, and Type~2 AUC.  Solid blue bars~= set
           size~20; hatched sky-blue bars~= set size~40.  Error bars:
           $\pm$1~SEM across $N = 200$ traces per model--condition cell.
           See Results for the Llama3.3:70b and Gemma3:12b dissociations
           under feedback.  Supplementary Figures~S7--S8 report
           Experiment~1 SDT; Figure~S9 reports Experiment~2 ROC
           curves.}
  \label{fig:sdt}
\end{figure}

Signal detection analyses corroborate and extend the behavioral
findings.  In Experiment~1, ceiling imagined-source accuracy fixed hit
rates near 1.0, so $d'$ indexes response conservatism rather than true
discrimination sensitivity, and negative criterion values ($c < 0$)
confirm a liberal bias toward internal responses; perceived-trial
Type~2 AUC nonetheless ranged from chance (0.49) to high (0.96--0.97)
across models and implementations (Supplementary Figures~S7--S8).

In Experiment~2 (Figure~\ref{fig:sdt}), SDT parameters were computed
per trace (200 per model $\times$ condition) and summarized as mean
$\pm$ SEM.  Without feedback most models show $d' \approx 0$,
confirming near-chance source discrimination despite above-chance mean
accuracy; Llama3.3:70b is a notable exception ($\bar{d}' = 2.01 \pm
0.11$ at set size~20, $1.62 \pm 0.11$ at set size~40).  Most striking
is a Type~1/Type~2 dissociation under feedback in Llama3.3:70b at set
size~20: $d'$ improves to $2.28 \pm 0.04$, yet Type~2 AUC, indexing
metacognitive accuracy \cite{galvin2003type}, declines below chance
($0.34 \pm 0.02$), so behavioral improvement under feedback is not
accompanied by improved calibration.  Gemma3:12b shows the inverse
pathology: $\bar{d}'$ turns negative under feedback ($-0.95 \pm 0.02$
at set size~20), reflecting systematic response inversion rather than
degraded discrimination, a qualitatively different failure mode.

\section{Discussion}

Source attribution in LLMs is not an invariant architectural property:
it is shaped by the structure of the conversational context.  That
single finding accounts for all main results: near-ceiling performance
when context-window retrieval was available, reversal to an
external-item advantage once episodic delay removed that shortcut, and
decoupling of accuracy from confidence under corrective feedback.
Scale did not drive this: Llama4:16x17b, the largest model in the
sample, performed comparably to smaller models across several measures
(Figures~\ref{fig:exp2} and~\ref{fig:sdt}), implicating individual
expert-network size within the mixture-of-experts architecture over
aggregate parameter count.

Accuracy and metacognitive calibration split in ways that depend on
architecture and condition, not overall capability.  In Experiment~1,
calibration held at the aggregate level in Single-Turn but broke down
across implementations: the same model could show positive calibration
in Single-Turn and reversed or undefined sensitivity in Trial-Chain.
Corrective feedback widened this split: Llama3.3:70b gained accuracy
while losing metacognitive sensitivity below chance; Gemma3:12b
inverted its source-discrimination mapping entirely.  Feedback does
not scale reality monitoring upward; it restructures how source
evidence maps to expressed confidence, direction depending on
architecture.

A separate failure operates before source attribution begins.  Reading
hallucinations, failures to reproduce the externally provided word,
are a mechanistically distinct upstream bottleneck that standard
human RM paradigms do not operationalize, since stimulus encoding is
experimenter- rather than participant-controlled.  In Experiment~1,
reading hallucinations were the strongest predictor of source
misattribution: when a model substitutes its own word for the
presented input, the downstream source judgment is ill-posed, and what
appears as a source-monitoring deficit is partly an encoding deficit
that must be separated out to draw valid inferences.  Models were not
entirely blind to this failure, however: reading hallucinations
independently reduced expressed confidence in Experiment~1, so some
metacognitive signal tracking encoding fidelity survives.  Trial-Chain
also reduced hallucinations for five of six models (the exception,
Gemma3:27b, rose from 0\% to 14.6\% while its quantization-aware
variant improved, a pattern absent in the 12B pair).  In Experiment~2,
reading hallucinations instead
escalated with feedback and larger set sizes and no longer predicted
confidence: a model that confidently misattributes a source it never
correctly encoded gives no signal to prompt correction, a failure
requiring external monitoring to detect.

These results extend Johnson and Raye's source monitoring
framework~\cite{johnson1981reality,johnson1988reality} to artificial
cognitive agents.  The framework holds that source judgments rest on
qualitative differences in attribute profiles: externally generated
items carry richer perceptual and contextual features; internally
generated items carry more cognitive operations and semantic
associations.  Consistent with this account, perceived items were
attributed substantially more accurately once episodic delay removed
the within-turn retrieval shortcut, driving ceiling performance on
imagined items in Experiment~1, exactly what the framework predicts
once perceptual and contextual attributes must carry the diagnostic
load across a temporal gap.  This advantage was largest without
feedback and narrowed under it, consistent with corrective signals
recalibrating imagined-item attribution rather than shifting only a
response criterion.

The human comparison is informative in both directions.  Human
reality-monitoring studies using closely related word-pair tasks
consistently report accuracy well above chance but far from ceiling,
with systematic externalizing biases and imperfect source
discrimination~\cite{ranjan2024reality,garrison2017monitoring}.
Several LLMs in Experiment~1 reached or approached ceiling, exceeding
the source-monitoring precision typically observed in these studies,
but fell substantially below that level in the Episodic-Chain (a
qualitative comparison; the human studies differ in design and trial
counts).  The metacognitive pattern also replicates a human finding:
expressed certainty was equally high for accurate and confabulated
source attributions on imagined items~\cite{ranjan2024reality},
suggesting that confidence--accuracy decoupling under memory load is
a broad feature of source attribution systems.

Feedback substantially increased reading hallucinations, most
strongly in larger conversational contexts, meaning that feedback
loops in LLM-mediated systems may amplify the hallucinations they aim
to reduce; this extends established findings that LLM performance
degrades with context length~\cite{liu2024lost} to encoding fidelity
under naturalistic multi-turn memory demands.  This is directly
relevant to agentic deployment, where high memory load and corrective
feedback co-occur by design: single-turn accuracy is not a reliable
proxy for multi-turn performance, models do not reliably self-correct
once off track, and encoding fidelity warrants monitoring as a
distinct safety metric alongside surface accuracy.

The $d'$/Type~2 AUC dissociation (source-discrimination accuracy
improving while metacognitive sensitivity collapses) raises a direct
concern about using expressed confidence as a trustworthiness signal:
when a model grows more accurate but its confidence becomes less
diagnostic of its own correctness~\cite{fleming2014measure}, downstream
systems relying on expressed certainty are misled.  Gemma3:12b's
response inversion adds a further caution: a model whose
source-attribution decisions are systematically reversed by corrective
signals cannot be treated as providing independent evidence about
source origins, regardless of confidence.  Both patterns are
compatible with the competing-biases account of Kumaran and
colleagues~\cite{kumaran2026competing}, in which LLM confidence
reflects a choice-supportive bias for self-generated answers and a
hypersensitivity to contradictory feedback, consistent with
Gemma3:12b's inversion taken to an extreme.  A parallel comes from
human calibration research: people are typically overconfident and
improve with practice only somewhat, unlike weather forecasters, who
check daily outcomes and are exceptionally well
calibrated~\cite{lichtenstein1982calibration}; the models here
received only a right/wrong label, uninformative about why an answer
was wrong, which may explain why feedback shifted confidence without
sharpening it.  LLM confidence should not be treated as a calibrated
probability estimate in high-stakes settings without independent
validation, consistent with broader evidence that LLM miscalibration
produces overconfident
errors~\cite{groot2024overconfidence,kadavath2022language}.  Existing
calibration benchmarks, which assess whether confidence tracks
accuracy on knowledge and reasoning
tasks~\cite{wang2026mirror,cacioli2026metacognitive,ackerman2026limited},
cannot detect this failure, since source memory imposes an additional
constraint beyond answer accuracy: which source an answer came from.
Deployment evaluations for multi-turn and agentic contexts should
therefore include source-memory assessments, since minimal-load
performance overestimates real-world capacity and scale alone does
not predict self-knowledge under memory load.

This study has four main limitations.  First, the word-pair paradigm,
adapted from human reality-monitoring research for direct
comparison~\cite{ranjan2024reality,johnson1981reality,johnson1988reality},
is deliberately minimal; generalization to richer deployment sources
(prior context turns, retrieved documents, tool outputs) and to less
related or factual content remains open.  Relatedness had at most a
negligible effect on accuracy (Experiment~2: OR $= 1.01$/unit;
Supplementary Information), but its upstream role reversed the human
pattern: in human memory, relatedness manufactures false memories,
since related completions come to mind effortlessly, leave sparse
process records, and are externalized at rates rising with the number
of related
items~\cite{ranjan2024reality,johnson1988reality,henkel1998reality,roediger1995creating};
here it instead aligned inputs with the model's own distributional
expectations, protecting verbatim reproduction (fewer reading
hallucinations, Experiment~1) and fluency-inflating confidence
independent of correctness (Experiment~2)~\cite{alter2009uniting}: a
source-confusion risk in humans but a copy-fidelity aid in LLMs,
directly testable by manipulating relatedness.  Second, the six models
tested span two families and two size classes with and without
quantization-aware training, but not the full range of current
architectures; all were run as instruction-tuned variants via Ollama,
so tuning and architectural effects are not separable, and
between-model estimates are specific to these six.  Whether
expert-network size, rather than aggregate parameter count, is the
operative scaling variable for source attribution is directly testable
across a broader range of mixture-of-experts models.  Third, all
analyses are behavioral; the mechanistic basis of these failure
modes, why feedback inverts source-response mappings in smaller
models and why confidence decouples from accuracy under memory load,
requires interpretability methods applied at the activation
level~\cite{conmy2023automated,zheng2025attention,bereska2024mechanistic},
for which these behavioral effects provide deployment-relevant
targets~\cite{basu2026interpretability,nanda2025pragmatic,kim2025agentic}.
Fourth, models were evaluated at a single decoding setting
(temperature fixed at 0, other parameters at framework defaults) and
a single feedback form; whether sampling temperature, presentation
order~\cite{pezeshkpour2023sensitivity}, or alternative feedback
modalities (delayed, partial, confidence-targeted) modulate these
effects is left to future work.

\section{Conclusion}

Source attribution in LLMs is not an invariant property of a model
but a function of how conversational memory is structured:
near-ceiling accuracy under single-trial demands, a reversal to an
external-item advantage under episodic delay, and a decoupling of
accuracy from confidence under corrective feedback.  This pattern
implicated individual expert-network size within mixture-of-experts
architectures over aggregate parameter count, suggesting scale alone
is the wrong lever for improving self-knowledge in deployed systems.
As AI systems shift from tools to autonomous agents operating across
extended interactions, evaluating what a model knows is not enough;
whether it can track where that knowledge came from is equally
essential, and reality monitoring offers a direct, model-comparable
behavioral test of that capacity.

\section{Methods}

\subsection{Large Language Models}

Six models of varying sizes were used from the Gemma~3 and Llama families,
all accessed as instruction-tuned variants via the Ollama inference
framework.  Model tags throughout (e.g., Gemma3:12b) follow Ollama's
naming convention, where the suffix after the colon denotes parameter
count and variant.  Models with 12 (Gemma3:12b) and 27 billion (Gemma3:27b)
parameters, including quantization-aware training variants
(Gemma3:12b-QAT, Gemma3:27b-QAT), were used from the Gemma~3 family.
From the Llama family, we used Llama~3.3 (70B; Llama3.3:70b) and
Llama~4 16$\times$17B (Llama4:16x17b), a sparse mixture-of-experts
model.  All models were run locally on a high-performance computing
cluster.  Decoding temperature was fixed at 0 (greedy decoding,
minimizing sampling variability across simulations); all other
inference parameters (top-p, context length) were left at Ollama
defaults for each model version.

\subsection{Stimuli}

Word-pair stimuli were drawn from a prior human RM paradigm~\cite{ranjan2024reality}
that sampled semantically related word pairs
from 36 Deese-Roediger-McDermott false-memory lists.  That prior study
established the human behavioral profile for this stimulus set:
participants showed externalizing biases (imagined items
misattributed as perceived) and source-dependent metacognitive
sensitivity, with calibration strongest for perceived items.
These shared materials enable direct human--LLM comparison.  Word pairs
were matched for relatedness and presented in randomized order.
Each second word was either externally provided (perceived source)
or generated by the model itself (imagined source).

\subsection{LLM Reality Monitoring Task}

All experimental interactions were administered using PsychScanner
(version~0.1), a purpose-built open-source Python package for running
structured behavioral experiments with large language models (source
code available at [BLINDED]).  Models received task instructions
specifying the RM procedure (word-pair source attribution and confidence
rating); no persona, role, or character framing was included beyond
the task description itself.  Trial instructions and stimuli were
serialized in JSON, a format that is both human- and machine-readable
and is well-represented in LLM pre-training corpora, minimizing the
risk that response format itself, rather than source-attribution capacity, drives model output.  Responses were parsed from structured
JSON output; trials that could not be parsed were logged as missing and
excluded from analysis.  To control for positional biases in JSON-structured prompts,
the order of response options in the source-attribution question was counterbalanced:
the default condition presented the \textit{external} option first
(\texttt{["external","internal"]}), and the counterbalanced condition reversed this
to \textit{internal} first (\texttt{["internal","external"]}).
This factor is included as a covariate (\texttt{order\_c}; dummy-coded: 0 =
external-first [reference], 1 = internal-first) in all models.
We implemented three distinct RM paradigms across two experiments, as described below.

\subsubsection{Experiment 1: Single-Trial Reality Monitoring}

In the single-trial RM task, the LLM completed each task independently,
with a fresh context window for each trial (or accumulated context in
the Trial-Chain condition).  For each trial, a word pair was given to
the LLM, which could be complete (perceived: both words provided) or
incomplete (imagined: first word provided, model generates second word).

In the \textit{Single-Turn} condition (Figure~\ref{fig:impl}a), the LLM
received all prompt instructions tailored to the RM task as a System
Message, and each trial was handled in a single interaction: the model
completed the task (word-pair completion, relatedness rating, source
monitoring, confidence rating) in one turn.

In the \textit{Trial-Chain} condition (Figure~\ref{fig:impl}b), the LLM
received sequential instructions across multiple turns while the trial
history was retained in the context window.  Each subtask (completion,
relatedness, source monitoring, confidence) was issued as a separate
turn.

A 2 (Source: Imagined vs.\ Perceived) $\times$ 2 (Memory Context:
Single-Turn vs.\ Trial-Chain) $\times$ 6 (Language Model) design was
used for Experiment~1.  Each cell comprised 144 trials (72 per source
type), for a total of 3,456 observations.

\subsubsection{Experiment 2: Episodic Reality Monitoring}

Experiment~2 extended this design with an episodic chain.
Reality monitoring trials were separated into an encoding phase (word-pair
completion and relatedness rating) and a delayed testing phase (source
attribution and confidence) over multiple trials within the same
conversational context.

Set size was manipulated to 20 and 40 interactions with LLMs, with equal
numbers across phases and sources (set size 20: 10 encoding $+$ 10
testing; set size 40: 20 encoding $+$ 20 testing).  Corrective feedback
was either present or absent throughout the episodic chain.  When present,
feedback had a uniform three-component structure across both phases,
delivered as a prefixed message at the start of the next trial's stimulus.
During encoding, the three components were: (1) a response-accuracy signal
indicating, for \textit{imagined} trials, whether the generated word was
novel (distinct from the first word in the pair and from any word produced in prior trials) or, for \textit{perceived} trials, whether the externally
provided word was correctly reproduced; (2) a rating-validity signal
confirming that the relatedness rating fell within the required 0--100\%
range; and (3) an overall trial-accuracy verdict, marked correct only if
both preceding components were satisfied.  During testing, the analogous
three components were: (1) a source-identification signal indicating whether
the model's origin attribution (\textit{imagined} vs.\ \textit{perceived})
was correct; (2) a rating-validity signal confirming that the confidence
rating fell within the required 1--6 range; and (3) an overall accuracy
verdict, correct only if both preceding components were satisfied.  This
design ensures that encoding-phase feedback targets generation or
reproduction fidelity depending on source type, while testing-phase feedback
targets source monitoring accuracy, keeping the two processes distinct
within the manipulation.
Each complete encoding-to-test sequence constituted a single
\textit{simulation trace} (hereafter \textit{trace}), the unit of
random-effects clustering.  This yielded a 2 (Set Size) $\times$
2 (Feedback) $\times$ 6 (Model) $\times$ 2 (Source) design with
4,800 total traces; source was a within-trace factor (both imagined
and perceived trials occurred within each trace), yielding 200~traces
per design cell.

\subsection{Experiment 1: Data Analysis}

All statistical analyses for Experiment~1 were conducted in Python
(version~3.11) using \texttt{statsmodels}.  The Experiment~1 dataset
contained 3,456 trials.

Trial compliance (binary) and recognition accuracy (binary) were modeled
using separate logistic regressions (binomial family, logit link; MLE)
on perceived trials only ($N = 1{,}728$); standard errors are
model-based (Fisher information).  One model (Gemma3:27b-QAT) exhibited
quasi-complete separation in the accuracy GLM (near-perfect
classification on perceived trials); its coefficient is reported but
flagged as unreliable and is not interpreted.  Relatedness ratings
(continuous) were analyzed using OLS regression with
heteroskedasticity-consistent (HC3) standard errors.  Type-II
likelihood-ratio tests (LRT) assessed significance of each predictor.
Confidence ratings (ordinal) were analyzed using ordered probit
regression.  Metacognitive sensitivity was operationalized as
Goodman--Kruskal $\gamma$ with Fisher $Z$ transformation ($\gamma_z$),
computed per model per memory condition.  Specific $\gamma$ values
cited in text are raw (untransformed); Figure~\ref{fig:exp1}d displays
Fisher $Z$-transformed $\gamma_z$.
Given the small sample ($n = 6$ models), only the memory factor could
be tested; formal tests are therefore exploratory.
No trials were excluded due to parse failures in Experiment~1 or
Experiment~2; responses were parsed successfully for all 3,456 and
72,000 trials, respectively.

To confirm that the primary inferences do not depend on fixed-effects
treatment of model identity, supplementary sensitivity analyses
re-estimated the reading hallucination, recognition accuracy, and
relatedness rating models with model as a random intercept
[\texttt{(1 | model)}] instead of a fixed categorical predictor.
All primary inferences replicated: the TrialChain advantage in
reading hallucinations ($z = -10.9$, $p < .001$), the detrimental
effect of reading hallucinations on accuracy ($z = -15.8$, $p < .001$),
and the source advantage in relatedness ratings ($t = 17.1$, $p < .001$).
Full random-effects sensitivity results are provided in the
Supplementary Information.

\subsection{Experiment 2: Data Analysis}

All statistical analyses for Experiment~2 were conducted in R
(version~4.5.1) via the \texttt{rpy2} interface in Python (version~3.11),
using the \texttt{lme4} and \texttt{lmerTest} packages for mixed-effects
models.  For the reading hallucination, recognition accuracy, and
metacognitive sensitivity ($\gamma$) outcomes, mixed models with random
effects for trace (simulation run) were used to account for clustering
of observations within simulation runs; signal-detection measures were
likewise computed per trace (see below).  Model fit was evaluated by likelihood-ratio tests
comparing nested model sequences (null $\rightarrow$ additive
$\rightarrow$ interactive), with the best-fitting model reported.
Fixed-effect significance was assessed via Type-II Wald $\chi^2$ tests
using the \texttt{car} package.  Marginal and conditional $R^2$ values
were computed using the \texttt{performance} package.  Estimated marginal
means and pairwise contrasts were computed using the \texttt{emmeans}
package.  All \texttt{lme4} GLMMs returned a formal convergence warning
(gradient unavailable via the default \texttt{bobyqa} optimizer); this
warning reflects the optimizer's inability to compute the analytical
gradient after convergence rather than a failure to converge, and is
common at large sample sizes where the Hessian is poorly conditioned.
Parameter estimates were consistent across two additional optimizers
(\texttt{Nelder\_Mead}, \texttt{nlminbwrap}), and are reported as
obtained.

Trial compliance (reading hallucinations; binary) on externally generated
trials ($N = 36{,}000$) was modeled using a GLMM (binomial family, logit
link) with the formula \texttt{read\_hallucination $\sim$ setsize
$\times$ fb\_exp + model + order\_c + (1 | trace)}.  Relatedness was
excluded from this model because reading hallucinations occur at the
word-reproduction stage before any relatedness-contingent source judgment
is formed; including relatedness would conflate a downstream covariate
with the outcome it precedes.  The interactive
model significantly improved fit over the additive model
(likelihood-ratio test, $\chi^2(1) =
20.7$, $p < .001$).

Recognition accuracy (binary) across all test-phase source-attribution
trials ($N = 72{,}000$; encoding-phase trials excluded) was modeled
using a GLMM (binomial family, logit link) with a random intercept for
trace, including mean-centered relatedness as a covariate.
The interactive model significantly improved fit over the
additive model ($\Delta\chi^2(3) = 2868.2$, $p < .001$;
$R^2_m = .150$, $R^2_c = .176$, where $R^2_m$ is the marginal $R^2$
attributable to fixed effects and $R^2_c$ is the conditional $R^2$
including random effects; Nakagawa and Schielzeth~\cite{nakagawa2013general}).  The small gap
between $R^2_m$ and $R^2_c$ (ICC $= .031$) indicates that, after
accounting for the fixed-effect predictors, between-trace variability
in accuracy is minor.  Full fixed-effect estimates for this model are
reported in the Supplementary Information.

Metacognitive sensitivity was operationalized as Fisher's
$Z$-transformed Goodman--Kruskal $\gamma$ computed within each
simulation, source condition, set size, and feedback condition
($N = 4{,}410$).  Exclusion rates for non-computable $\gamma$ varied
significantly by model, set size, feedback, and source (chi-square
test of independence, $\chi^2(47) = 3489.1$, $p < .001$) but were not
concentrated in a single condition: the only fully deterministic cell
(Gemma3:27b-QAT, no feedback, imagined source; 400 of 400 traces
excluded) accounted for 7.7\% of total exclusions, with several other
model--condition cells (e.g., Gemma3:27b under feedback, Gemma3:12b
without feedback) each contributing a comparably large share.
Gamma coefficients were analyzed using an LMM with
random intercepts only (\texttt{(1 | trace)}); random slopes for source
were structurally unidentifiable because each trace contributed only
two observations per source condition, preventing reliable estimation
of within-trace variation in source-specific slopes.  The intercept-only
random effect yielded a singular fit (estimated between-trace variance
$\approx 0$), indicating that the between-trace variance in mean
$\gamma_z$ is negligible after accounting for fixed effects.  When
between-trace variance is zero, the mixed model reduces to OLS; we
confirmed this by re-estimating all fixed effects under OLS with
heteroskedasticity-consistent (HC3) standard errors, obtaining
point estimates within rounding of the LMM values and CIs of
comparable width, validating the LMM fixed-effect inferences.  The
interactive model significantly improved fit
($\Delta\chi^2(3) = 355.6$, $p < .001$; AIC: 24{,}436.8 interactive vs.\ 24{,}782.6 additive;
$R^2_m = .29$).

Confidence ratings (ordinal, 1--6) were analyzed using a Cumulative
Link Model (CLM; logit link, flexible thresholds;
Christensen~\cite{christensen2023ordinal}) via R's \texttt{ordinal} package.
Unlike the other Experiment~2 outcomes, this model does not include a
random effect for trace: a Cumulative Link Mixed Model with
\texttt{(1 | trace)} exhibited quasi-complete separation on model
identity (Llama4:16x17b had 0/12{,}000 trials at confidence~$=6$ and
Gemma3:27b-QAT had 3/12{,}000 at confidence~$=1$), producing a
non-positive-definite Hessian for which confidence intervals and
omnibus tests could not be computed. Confidence is therefore reported
as a secondary, descriptive outcome using the fixed-effects CLM.
To confirm that the trial-level accuracy null does not depend on this
specification, the effect was re-tested at the trace level, where
observations are independent: within each trace containing both
correct and incorrect trials ($N = 4{,}519$ of 4{,}800; the remainder,
predominantly all-correct Llama3.3:70b traces, had invariant accuracy
and contribute no within-trace contrast), mean confidence on correct
trials was contrasted with mean confidence on incorrect trials.  The
paired difference was null overall ($M = +0.001$ points on the 1--6
scale, 95\% CI $[-0.011,\, +0.013]$, $t(4518) = 0.11$, $p = .91$,
Cohen's $d_z = 0.002$), with small opposite-signed effects by feedback
condition ($+0.06$ without feedback, $-0.06$ with feedback, both $p < .001$), consistent with feedback reversing, rather than merely
weakening, the confidence--accuracy association, in line with the
metacognitive-sensitivity and Type~2 AUC results.
Fixed effects included source, setsize, fb\_exp, three two-way
interactions (source $\times$ setsize, source $\times$ feedback,
setsize $\times$ feedback), model, accuracy (correct vs.\ incorrect
source judgment), reading\_hallucination, mean-centered relatedness
(rating\_cen), and response-option order (order\_c).  An interactive
model was compared to additive and null specifications by
likelihood-ratio tests (LRT).  Omnibus significance of each predictor
was assessed using the \texttt{car::Anova} function with Type-II tests;
for CLM models this function computes likelihood-ratio tests (profile
deviance) rather than Wald $\chi^2$ approximations, so the
predictor-level $\chi^2$ values reported in Results reflect LRT
statistics and may differ from the squared Wald $z$ values in the
coefficient table.  Post-hoc pairwise contrasts for source $\times$
feedback and source $\times$ setsize interactions used Bonferroni
correction (\texttt{emmeans} package).  Model-comparison statistics
and full odds-ratio tables are reported in the Supplementary Information.

Cumulative accuracy trajectories across test trials (Supplementary
Figure~S6) were examined descriptively.  Complementing this, OLS
regression with heteroskedasticity-consistent (HC3) standard errors
($R^2 = .21$, $n = 2{,}183$ simulations with computable $\gamma$)
predicted Fisher-Z--transformed metacognitive $\gamma$ on perceived
trials from set size, feedback, model identity, set size $\times$
feedback interaction, and mean reading hallucination rate as a covariate.

To confirm that the primary Experiment~2 inferences do not depend on
fixed-effects treatment of model identity, supplementary sensitivity
analyses re-fitted the reading hallucination GLMM, recognition
accuracy GLMM, and metacognitive $\gamma_z$ LMM with model added as
a crossed random intercept [\texttt{(1 | model) + (1 | trace)}].
All primary fixed-effect inferences replicated in direction and
significance: the setsize~$\times$~feedback interaction on reading
hallucinations ($\chi^2(1) = 21.9$, $p < .001$), the
source~$\times$~feedback interaction on accuracy ($\chi^2(1) =
2{,}734.7$, $p < .001$), and all six terms of the $\gamma_z$ model
remained significant.  Full sensitivity results are provided in the Supplementary
Information.

All VIF values were below 5.  Odds ratios with 95\% CIs are reported
for binary outcomes; unstandardized coefficients with 95\% CIs for
continuous outcomes.  All $p$-values are two-tailed.


\section*{Code and Data Availability}

Analysis code, stimulus files, and anonymized experimental data are
openly available at \url{https://github.com/saurabhr/LLM-RM/} (MIT
license).  The PsychScanner stimulus-presentation package (version~0.1)
used to administer all LLM interactions is available at
\url{https://github.com/saurabhr/psychscanner_v_0_1_0}.


\section*{Declaration of Generative AI and AI-Assisted Technologies in the Writing Process}

The authors used Claude (Anthropic, version claude-sonnet-4-6) during the preparation of this
manuscript. The tool was used in the paper text for language editing, clarity improvements,
\LaTeX{} typesetting, and code development. The authors reviewed, edited, and validated all AI-assisted outputs and
made all core intellectual and design decisions.


\backmatter

%
%
%

\bmhead{Acknowledgements}
S.R.\ was supported by the Threadgill Dissertation Fellowship,
University of Florida.

\bmhead{Author contributions}
S.R.: Conceptualization, Visualization, Software, Methodology,
  Formal analysis, Data curation, Writing (original draft and revisions).
  K.S.: Conceptualization, Writing (review \& editing).
  B.O.: Supervision, Writing (review \& editing).

\bmhead{Competing interests}
None declared.

\bmhead{ORCID}
Saurabh Ranjan: 0000-0002-7868-7223\\
Konstantina Sokratous: 0000-0003-4489-5494\\
Brian Odegaard: 0000-0002-5459-1884



\clearpage
\section*{Supplementary Information}

\renewcommand{\thefigure}{S\the\numexpr\value{figure}-6\relax}
\renewcommand{\thetable}{S\arabic{table}}
\setcounter{table}{0}


\section*{Experiment 1 Supplemental Figures}

\begin{figure}[htbp]
  \centering
  \includegraphics[width=\textwidth]{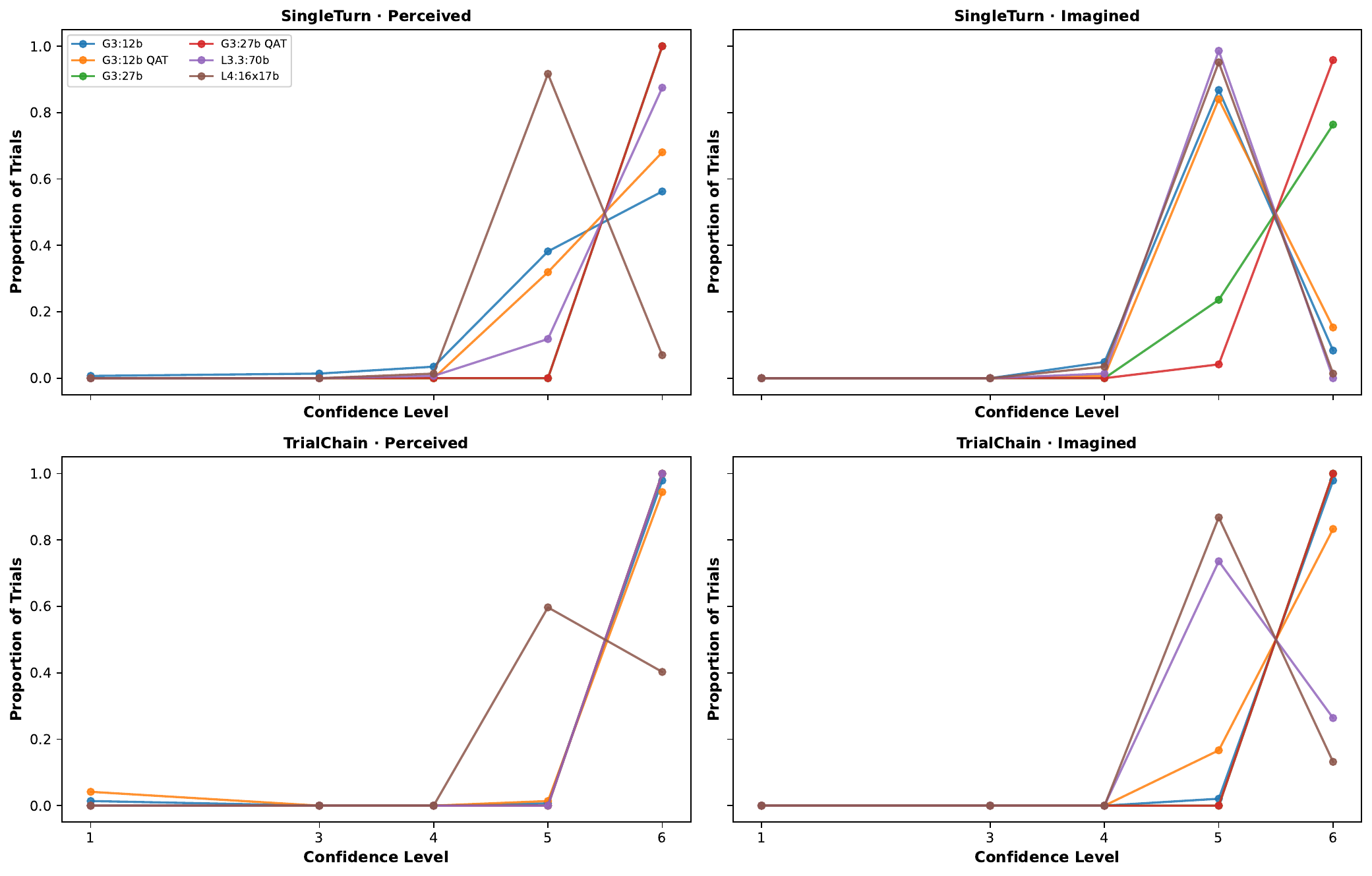}
  \caption{%
    \textbf{Imagined-item confidence is concentrated at the highest
    confidence levels across all models, reflecting ceiling accuracy;
    perceived-item distributions show substantial between-model
    variability.}
    Proportion of trials at each confidence level (1~= low certainty;
    6~= high certainty) for Single-Turn (top row) and Trial-Chain
    (bottom row) conditions, separated by source type (external: left
    column; internal: right column) and model (one line per model,
    color coding as in Figure~3 of the main text).
    Imagined (internal) item distributions are steeply left-skewed (concentrated at confidence levels 5--6, with a thin tail toward low confidence) across all models and both implementations.  This pattern follows directly from
    ceiling-level accuracy on imagined trials: because the generated
    word remains in the active context at the time of source
    attribution, models identify the source correctly and with high
    certainty.
    Perceived (external) item distributions show considerably more
    between-model variability, with models experiencing higher
    hallucination rates producing flatter, more uniformly distributed
    confidence profiles consistent with genuine source uncertainty.
    No substantial difference in distributional shape is evident between
    Single-Turn and Trial-Chain, suggesting that decomposing subtasks
    across turns does not alter overall confidence calibration.
  }
  \label{fig:s1}
\end{figure}

\clearpage

\begin{figure}[htbp]
  \centering
  \includegraphics[width=\textwidth]{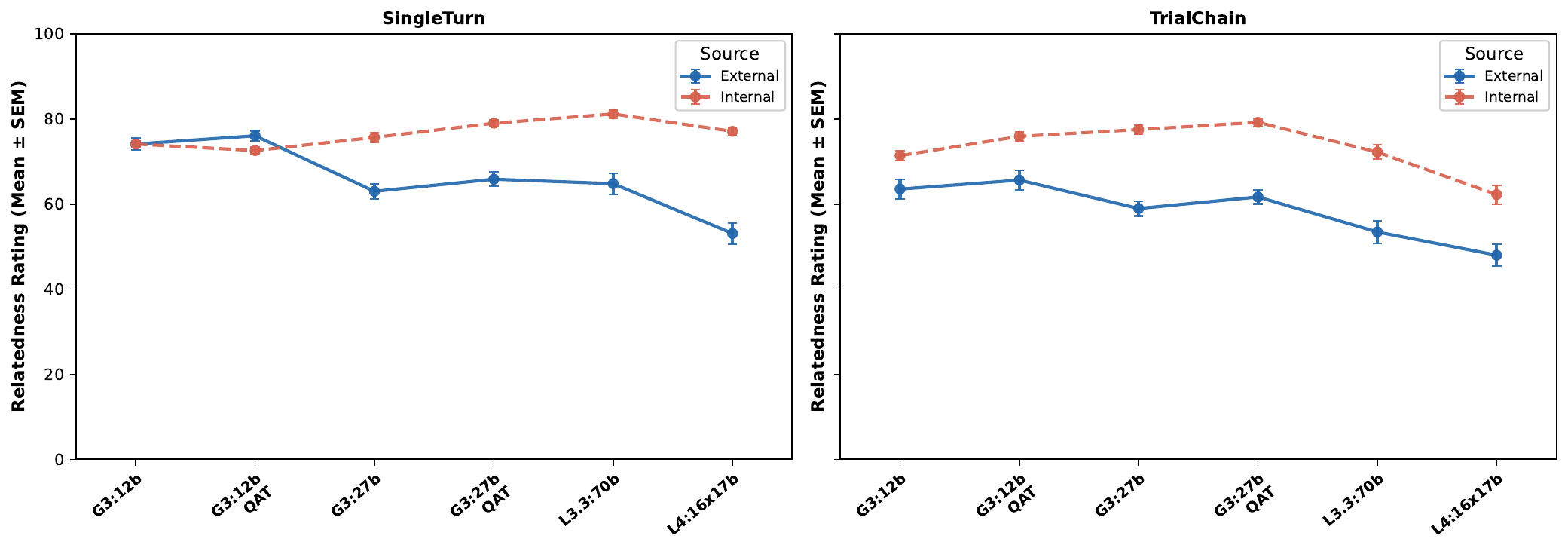}
  \caption{%
    \textbf{Imagined word pairs are rated more semantically related
    than perceived pairs; ratings are otherwise broadly stable across
    models and memory implementations.}
    Mean word-pair relatedness ratings (0~= unrelated, 100~= highly
    related; $\pm$1~SEM across trials within each model--condition cell)
    for Single-Turn (left) and Trial-Chain (right) conditions, by
    source type (external: blue solid; internal: red dashed) and model.
    Source type was the dominant predictor of relatedness ratings
    (Table~\ref{tab:s2_rr_anova}): imagined (internal) pairs were rated
    as more related than perceived (external) pairs in most
    model--memory cells (10 of 12), by 12.8 points on average (Type-II
    ANOVA, partial $\eta^2 = .078$, the largest effect in the model).
    This is consistent with imagined completions being the model's own
    most strongly associated continuation of the cue word, whereas
    perceived (experimenter-supplied) pairs were not selected to
    maximize associative strength.
    No model showed a substantial change in relatedness ratings between
    Single-Turn and Trial-Chain, indicating that the task decomposition
    manipulation did not alter how models perceived the semantic
    structure of the stimuli.
    Relatedness was included as a covariate in all generalized linear
    models predicting source accuracy; its effect was small relative
    to the primary predictors (reading hallucination and model
    identity) and was not statistically reliable in the Experiment~1
    accuracy model ($b = -0.003$, $p = .37$).
  }
  \label{fig:s2}
\end{figure}

\clearpage

\begin{figure}[htbp]
  \centering
  \includegraphics[width=\textwidth]{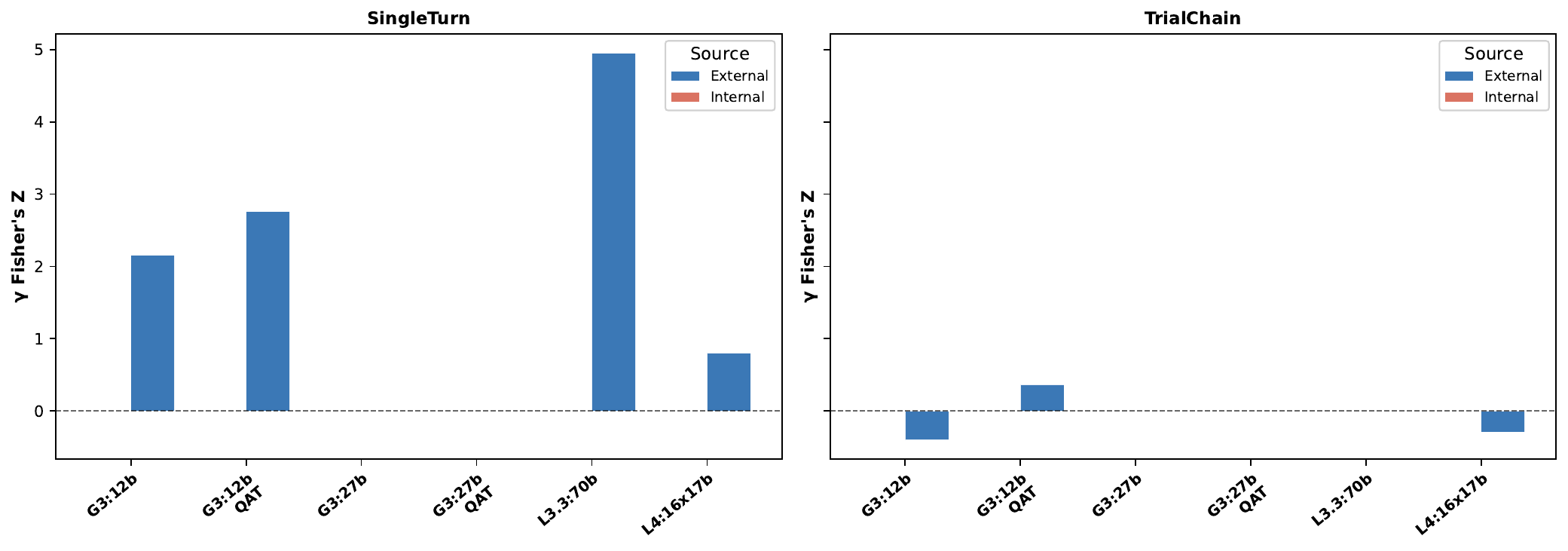}
  \caption{%
    \textbf{Perceived-source metacognitive sensitivity is positive in
    most cases where computable; imagined-source $\gamma$ is undefined
    across most models due to ceiling accuracy.}
    Goodman--Kruskal $\gamma$ coefficients (Fisher $Z$-transformed)
    relating expressed confidence to trial-level source accuracy for
    Single-Turn (left) and Trial-Chain (right) conditions, shown
    separately for external (blue) and internal (orange) sources.
    Positive $\gamma$ indicates above-chance metacognitive sensitivity
    (higher confidence on correct than incorrect trials); negative
    values indicate below-chance calibration.  $\gamma$ was computed
    once per model--condition--source cell from all trials in that
    cell pooled.
    Imagined (internal) item bars are absent for most models because
    ceiling-level accuracy (all or nearly all trials correctly attributed) yields zero variance in the correctness variable,
    making $\gamma$ mathematically undefined.  The absence of
    imagined-source bars is a result, not a data gap: it reflects a
    paradigm in which the source signal is too transparent to generate
    the outcome variance needed for calibration measurement.
    Where perceived-source $\gamma$ is computable, it is positive in
    most cases (5 of 7 model--condition cells), indicating above-chance
    calibration on externally provided items; the two exceptions
    (Gemma3:12b and Llama4:16x17b, both under Trial-Chain) show negative
    $\gamma$, indicating below-chance calibration in those cells.
    Single-Turn showed higher $\gamma$ in most
    computable cases than Trial-Chain, consistent with the simpler
    within-turn retrieval context supporting marginally better
    metacognitive calibration.
  }
  \label{fig:s3}
\end{figure}

\clearpage


\section*{Experiment 2 Supplemental Figures}

\begin{figure}[htbp]
  \centering
  \includegraphics[width=\textwidth]{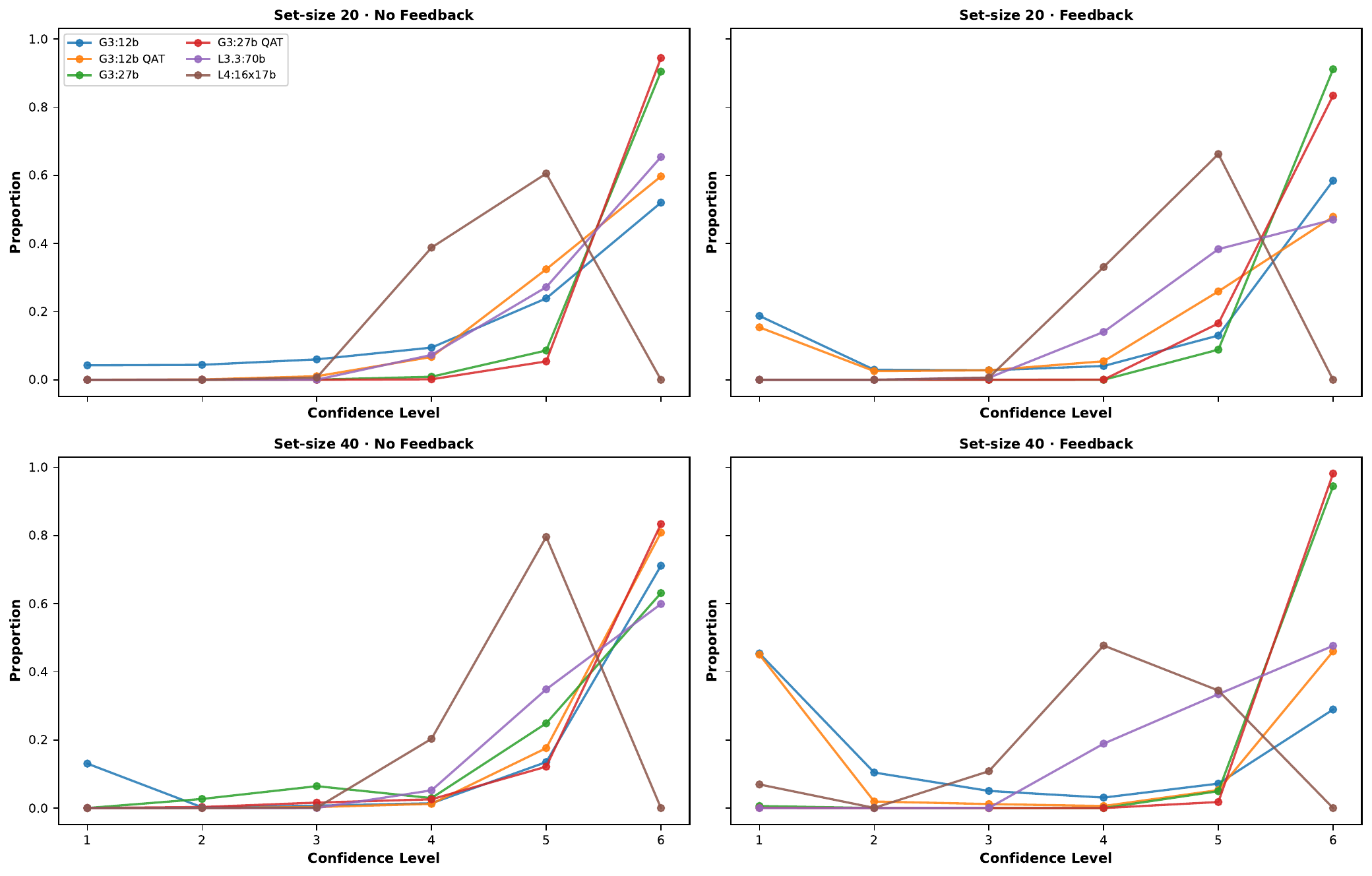}
  \caption{%
    \textbf{Corrective feedback reduces expressed confidence after
    controlling for model identity; larger set size amplifies this
    suppression.}
    Proportion of trials at each confidence level (1~= low; 6~= high)
    across the four Episodic-Chain conditions: top row~= set size~20;
    bottom row~= set size~40; left column~= no feedback; right column~=
    feedback.  Each panel shows one line per model across all source
    types combined.
    Without feedback, distributions are broadly spread across the 1--6
    range in most models, reflecting genuine uncertainty about source
    attributions under episodic delay.
    Despite apparent visual differences across panels, the Cumulative
    Link Model (CLM; main text) shows that corrective feedback
    significantly \textit{reduced} expressed confidence after
    controlling for model identity, source type, and set size
    ($b = -0.15$, $\text{OR} = 0.86$, $p < .001$).
    Trial-level accuracy was not a significant predictor of confidence
    in the CLM after controlling for other factors, indicating that
    the confidence shift under feedback is decoupled from item-level
    correctness rather than tracking it.
    The set-size~$\times$~feedback interaction
    ($\chi^2(1) = 306.6$, $p < .001$; $\text{OR} = 0.56$) shows
    that feedback suppressed confidence most sharply at set size~40,
    consistent with greater uncertainty under longer encoding lists.
  }
  \label{fig:s4}
\end{figure}

\clearpage

\begin{figure}[htbp]
  \centering
  \includegraphics[width=\textwidth]{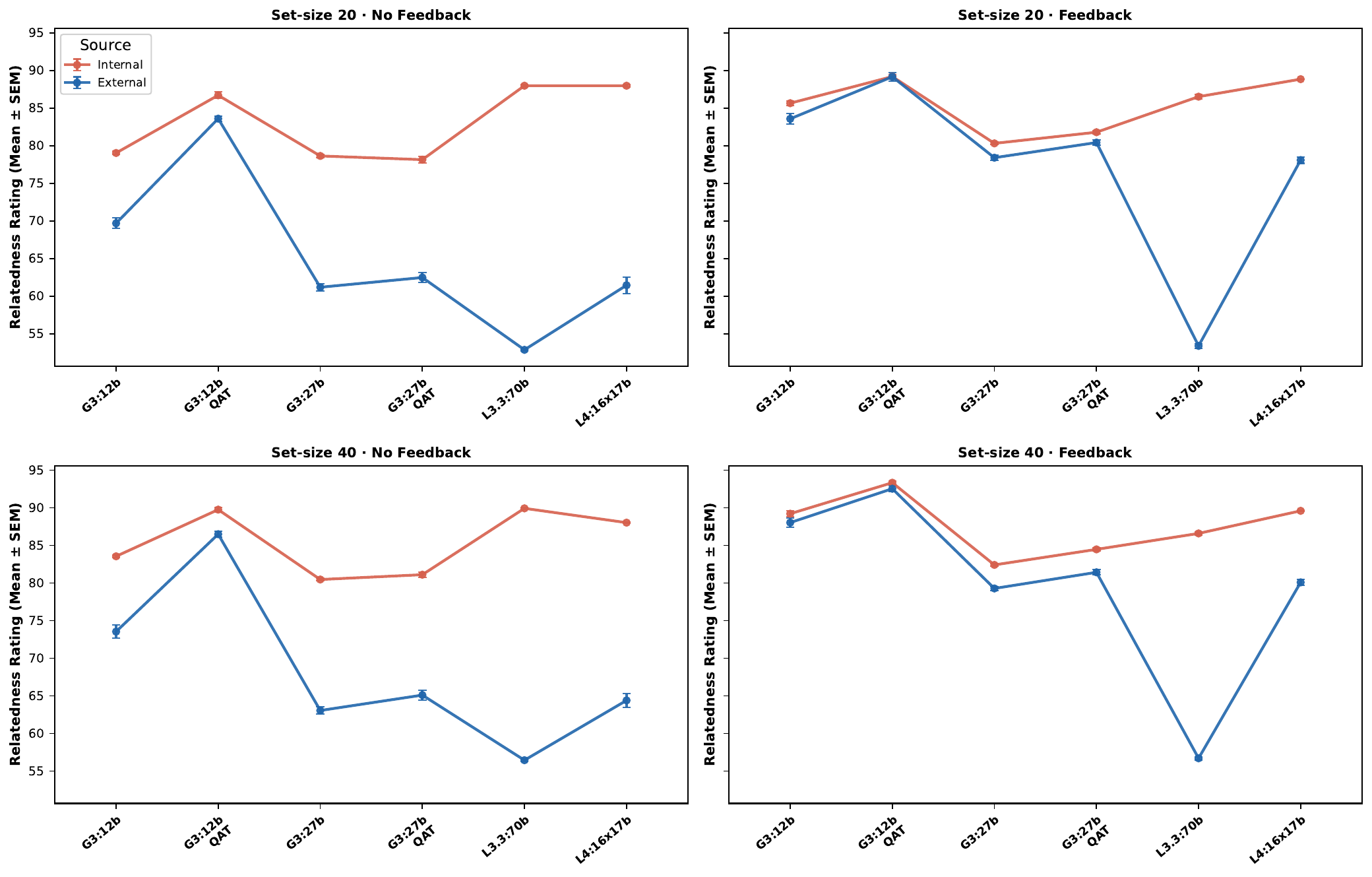}
  \caption{%
    \textbf{Imagined word pairs are rated more related than perceived
    pairs, closely replicating Experiment~1; this gap narrows under
    feedback but is stable across set size.  Higher relatedness
    independently predicts slightly higher expressed confidence.}
    Mean word-pair relatedness ratings (0~= unrelated, 100~= highly
    related; $\pm$1~SEM across trace IDs, $N = 200$ per cell, within
    each model--condition cell) across the four Episodic-Chain
    conditions (top row: set size~20; bottom row: set size~40; left
    column: no feedback; right column: feedback), by source type
    (internal: red; external: blue) and model.
    Source type was again the dominant predictor of relatedness
    (linear mixed model, Type-II Wald $\chi^2(1) = 8397.7$, $p < .001$):
    imagined pairs were rated 12.8 points higher than perceived pairs
    on average (emmeans contrast, perceived $-$ imagined $= -12.82$,
    $SE = 0.14$, $z = -91.6$, $p < .001$), closely replicating, not attenuating, the 12.8-point gap observed in Experiment~1
    (Figure~\ref{fig:s2}).
    This gap interacted with feedback (source~$\times$~feedback,
    $\chi^2(1) = 1286.9$, $p < .001$): feedback was associated with a
    substantially larger \textit{increase} in perceived-pair ratings
    ($+11.71$ points, $p < .001$) than in imagined-pair ratings
    ($+2.24$ points, $p < .001$), narrowing but not eliminating the
    source-type gap under feedback.  There was no significant
    source~$\times$~set-size interaction
    ($\chi^2(1) = 2.3$, $p = .128$): the relatedness gap was stable
    across set sizes.
    Despite this structure, the absolute range of variation was
    narrow, and higher relatedness remained a reliable predictor of
    marginally higher expressed confidence in the Episodic-Chain CLM
    (OR~= 1.01 per unit on the 0--100 scale; main text), independent
    of source type, feedback condition, set size, and model identity.
  }
  \label{fig:s5}
\end{figure}

\clearpage


\section*{Experiment 2 Cumulative Accuracy Trajectories}

\begin{figure}[htbp]
  \centering
  \includegraphics[width=\textwidth]{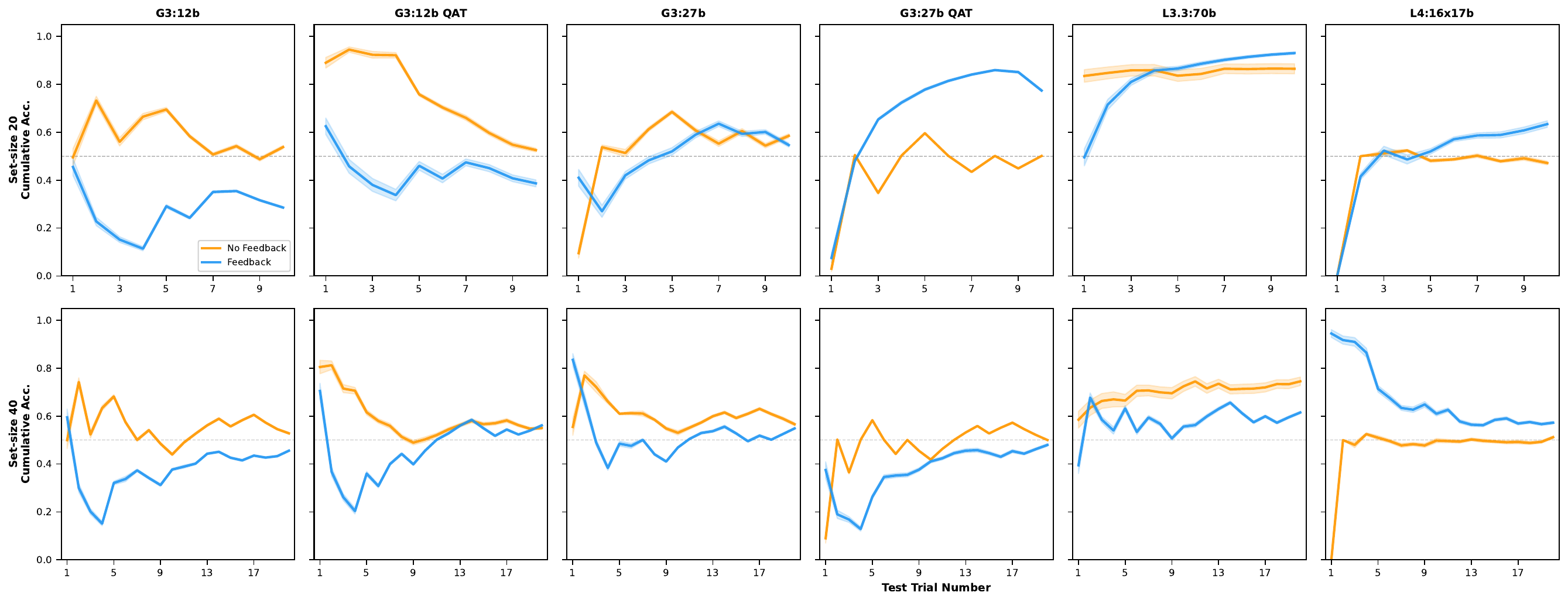}
  \caption{%
    \textbf{Cumulative accuracy trajectories are heterogeneous across
    models and do not follow a single scale- or feedback-dependent
    pattern.}
    Cumulative mean accuracy across successive test trials in the
    Episodic-Chain paradigm (Experiment~2).  Each column represents
    one model (left to right: Gemma3:12b, Gemma3:12b-QAT, Gemma3:27b,
    Gemma3:27b-QAT, Llama3.3:70b, Llama4:16x17b); rows correspond to
    set size (set size~20: top row; set size~40: bottom row).  Orange
    lines: no-feedback condition; blue lines: feedback condition.
    Shaded bands: $\pm$1~SEM across $N = 200$ traces per
    model--condition cell.
    Trajectories varied substantially across models and set sizes,
    with no single directional pattern dominating either feedback
    condition.  Without feedback, Gemma3:12b-QAT showed the clearest
    and most consistent decline across the test sequence at both set
    sizes; Gemma3:27b-QAT and Llama4:16x17b instead rose over the
    sequence at both set sizes (the latter plateauing after an early
    rise); Llama3.3:70b remained comparatively high throughout with
    little or no decline; Gemma3:12b showed a weak downward trend with
    substantial fluctuation at both set sizes; and Gemma3:27b showed
    opposite trends at the two set sizes (rising at set size~20,
    declining at set size~40).  Under feedback, Gemma3:12b and
    Gemma3:12b-QAT showed an initial decline followed by partial
    recovery at both set sizes, and Gemma3:27b-QAT improved
    consistently at both set sizes despite reaching a substantially
    lower ceiling at set size~40.  Trajectory shape depended jointly
    on model identity and set size rather than on model scale alone:
    Llama4:16x17b, the largest model in the sample, showed the
    most extreme reversal in the figure, improving steadily at set
    size~20 but declining sharply at set size~40, and Llama3.3:70b
    showed a clear improvement at set size~20 that became weak and
    noisy at set size~40.
    These trial-level trajectories are descriptive; the statistically
    tested source~$\times$~feedback and source~$\times$~setsize
    interactions are reported via the GLMMs in the main text
    (Figure~4), and individual model-level trajectory features here
    should be interpreted cautiously given the modest number of trials
    per condition.
  }
  \label{fig:s6}
\end{figure}

\clearpage


\section*{Signal Detection Theory — Experiment 1}

SDT measures were computed using the Python packages \textit{scikit-learn} \cite{sklearn2011} and \textit{SciPy}.
Hit and false-alarm counts were extracted from the confusion matrix;
AUC and ROC curves were derived from signed confidence ratings;
and the inverse normal transformation supplied the $z$-score transformation yielding $d'$ and criterion $c$.
Signal trials were \textit{imagined} (internally generated) word pairs;
noise trials were \textit{perceived} (externally provided) word pairs.
A hit was scored when the model correctly identified an imagined pair as internal;
a false alarm was scored when a perceived pair was incorrectly called internal.
Log-linear correction \cite{hautus1995corrections} was applied to prevent
infinite $z$-scores when hit rate or false-alarm rate reached 0 or 1.

\begin{figure}[htbp]
  \centering
  \includegraphics[width=\textwidth]{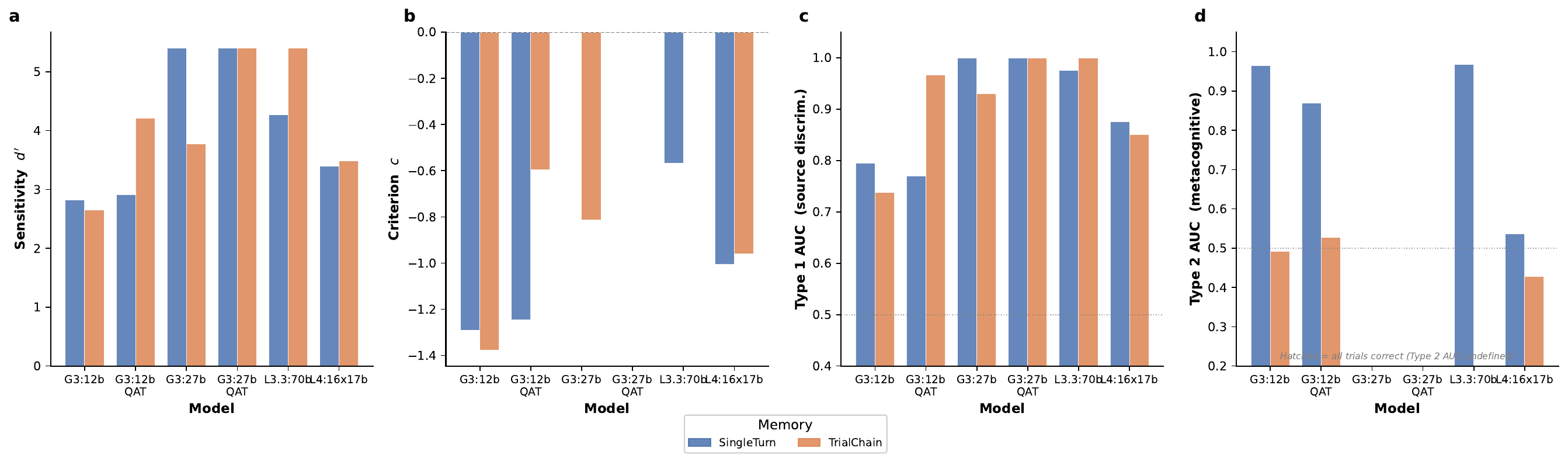}
  \caption{%
    \textbf{Experiment~1 SDT metrics are interpretively constrained
    by hit-rate ceiling: $d'$ reflects response conservatism rather
    than sensitivity, and Type~2 AUC is restricted to perceived
    trials.}
    Four panels (left to right): $d'$ (Type~1 sensitivity), criterion
    $c$ (response bias), Type~1 AUC (source discrimination via signed
    confidence ratings), and Type~2 AUC (metacognitive sensitivity).
    Grouped bars show Single-Turn (blue) and Trial-Chain (orange)
    values for each model.  Hatched bars in the Type~2 AUC panel
    indicate cells where the metric was undefined because perceived
    accuracy also reached ceiling, leaving no accuracy variance.
    Because all models achieved near-perfect accuracy on imagined
    items (hit rate~$\approx 1.0$ after log-linear correction),
    $d'$ variation across models reflects false-alarm rate differences only, not sensitivity.  A model with a low false-alarm rate
    appears to have high $d'$ not because it discriminates sources
    better but because it rarely calls a perceived item ``imagined.''
    Criterion $c$ values are negative in 8 of 12 model--condition cells,
    confirming a liberal response bias toward the internal attribution
    in most models; the remaining 4 cells, the same near-ceiling cells noted above (Gemma3:27b Single-Turn, Gemma3:27b-QAT Single-Turn, Gemma3:27b-QAT TrialChain, Llama3.3:70b TrialChain), show $c = 0$, reflecting an unbiased rather than a liberal
    response criterion.
    Type~2 AUC was computed restricted to perceived trials to avoid
    conflating source-level confidence differences (imagined items receive reliably lower confidence than perceived items) with
    metacognitive calibration; including imagined trials would inflate
    AUC by confounding source-accuracy level with within-source
    calibration.  Type~1 AUC was computed via signed confidence
    ratings (internal judgment~= positive score; external~= negative score).
    Log-linear correction applied~\protect\cite{hautus1995corrections}.
  }
  \label{fig:s7}
\end{figure}

\textbf{Interpretive caveat: hit-rate ceiling.}
All models achieved imagined-source accuracy of 1.00 in Experiment~1, fixing
$HR \approx 0.997$ across all cells after log-linear correction.
Because $d' = z(HR) - z(\text{FA})$ and $c = -0.5[z(HR) + z(\text{FA})]$ are
both functions of the same fixed $z(HR) \approx 2.75$, neither metric can
distinguish sensitivity from response bias in this experiment.
$d'$ values in Figure~\ref{fig:s7} reflect \textit{false-alarm rate variation only}
and should be interpreted accordingly.

\textbf{Type 1 and Type 2 AUC.}
Two AUC measures are reported throughout.
\textit{Type 1 AUC} indexes source discrimination:
it is the area under the ROC curve obtained by
ranking trials by a signed confidence score
(confidence rating multiplied by $+1$ for internal judgments and $-1$ for external judgments),
and discriminating imagined from perceived trials.
Type 1 AUC $= 0.5$ indicates chance-level source discrimination;
Type 1 AUC $= 1.0$ indicates perfect discrimination.
\textit{Type 2 AUC} indexes metacognitive sensitivity:
it is the area under the ROC curve obtained by
ranking trials by raw confidence rating and discriminating
\textit{correct} from \textit{incorrect} source judgments
\cite{galvin2003type}.
Type 2 AUC $= 0.5$ means confidence is unrelated to accuracy
(no metacognitive sensitivity); Type 2 AUC $> 0.5$ means
higher confidence reliably predicts correct responses.
Type 2 AUC is undefined (reported as NaN) when all trials
in a cell were correct (no variance in accuracy to discriminate)
or when confidence ratings are constant across all trials in a cell
(no variance in confidence).

\textbf{Type 2 AUC: perceived-trial restriction (Experiment~1).}
Because imagined accuracy is 1.00 for all Experiment~1 cells, an all-trials
Type~2 AUC would be confounded: imagined trials always enter the ``correct''
class yet carry systematically \textit{lower} raw confidence than perceived
trials (marginal means 5.51 vs.\ 5.77 on the 1--6 scale; consistent in
direction with the ordered-probit source effect in the main text), so
source-level confidence differences would contaminate the
correct-versus-incorrect discrimination that Type~2 AUC is meant to index.
Type~2 AUC in Experiment~1 is therefore computed restricted to \textit{perceived
trials only}, providing an unconfounded calibration estimate.
For cells where perceived accuracy also reaches ceiling
(Gemma3:27b Single-Turn, Gemma3:27b-QAT Single-Turn, Gemma3:27b-QAT
TrialChain, Llama3.3:70b TrialChain) or where confidence ratings are
constant across all perceived trials (Gemma3:27b TrialChain), Type~2
AUC remains undefined (NaN).

\clearpage

\begin{figure}[htbp]
  \centering
  \includegraphics[width=\textwidth]{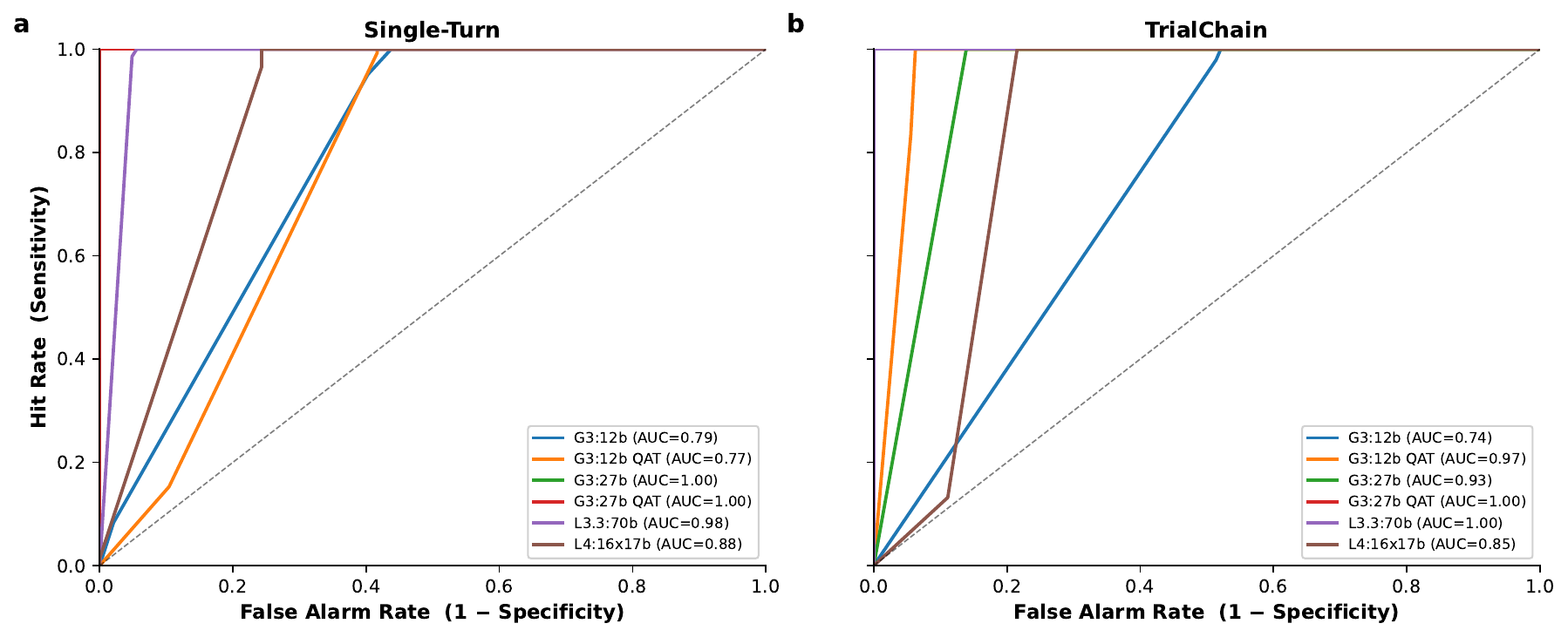}
  \caption{%
    \textbf{Near-ceiling hit rates compress all Experiment~1 ROC
    curves to the top-left corner; between-model AUC differences
    reflect false-alarm rate variation rather than differential
    discrimination sensitivity.}
    ROC curves for each model in Single-Turn (top row) and
    Trial-Chain (bottom row) implementations, showing hit rate
    (sensitivity) as a function of false-alarm rate
    (1~$-$~specificity) across confidence thresholds.  Curves
    computed from signed confidence ratings
    (internal judgment~= positive score; external~= negative score).
    Dashed diagonal~= chance discrimination (AUC~= 0.50).  AUC per
    model reported in the legend.
    Because imagined-item hit rates are at or near 1.0 for all
    models, all curves originate from the top-left corner of ROC
    space regardless of model.  The only dimension of variation is
    in false-alarm rate (how frequently models call a perceived item ``imagined''), which shifts curves leftward (lower
    false-alarm rate, higher apparent AUC) or rightward (higher
    false-alarm rate, lower AUC).  Between-model AUC differences
    here are therefore driven entirely by response criterion
    variation, not genuine differences in source-discrimination
    ability.  Experiment~2 ROC curves (Figure~\ref{fig:s9}) do not
    share this constraint: episodic delay removes the hit-rate
    ceiling and allows ROC curves to vary in both dimensions,
    permitting interpretation of AUC as a measure of discrimination
    sensitivity rather than response bias.
  }
  \label{fig:s8}
\end{figure}

\clearpage


\section*{Signal Detection Theory --- Experiment 2}

SDT measures for Experiment 2 were computed per \textit{trace}, treating each trace
as an independent participant ($N = 200$ traces per model $\times$ condition cell).
Pooling trials across traces, as is common in machine-learning evaluations, does not yield the mean of per-trace $d'$ values and precludes valid uncertainty
quantification \cite{macmillan2005detection}.
Per-trace $d'$, $c$, Type~1 AUC, and Type~2 AUC were computed using the same
helper functions as Experiment~1 (see above).
Set size~20 cells contain 5 imagined $+$ 5 perceived trials per trace;
set size~40 cells contain 10 imagined $+$ 10 perceived.
AUC values are reported as means excluding traces for which the metric was
undefined (constant score or constant accuracy).
Figure~6 (main text) reports mean values across 200 traces per cell with
error bars of $\pm$1 SEM.
Conditions are defined by set size crossed with feedback (no-feedback vs.\ feedback):
set size 20 (10 encoding $+$ 10 test trials per trace) and set size 40
(20 encoding $+$ 20 test trials per trace).


\begin{figure}[htbp]
  \centering
  \includegraphics[width=\textwidth]{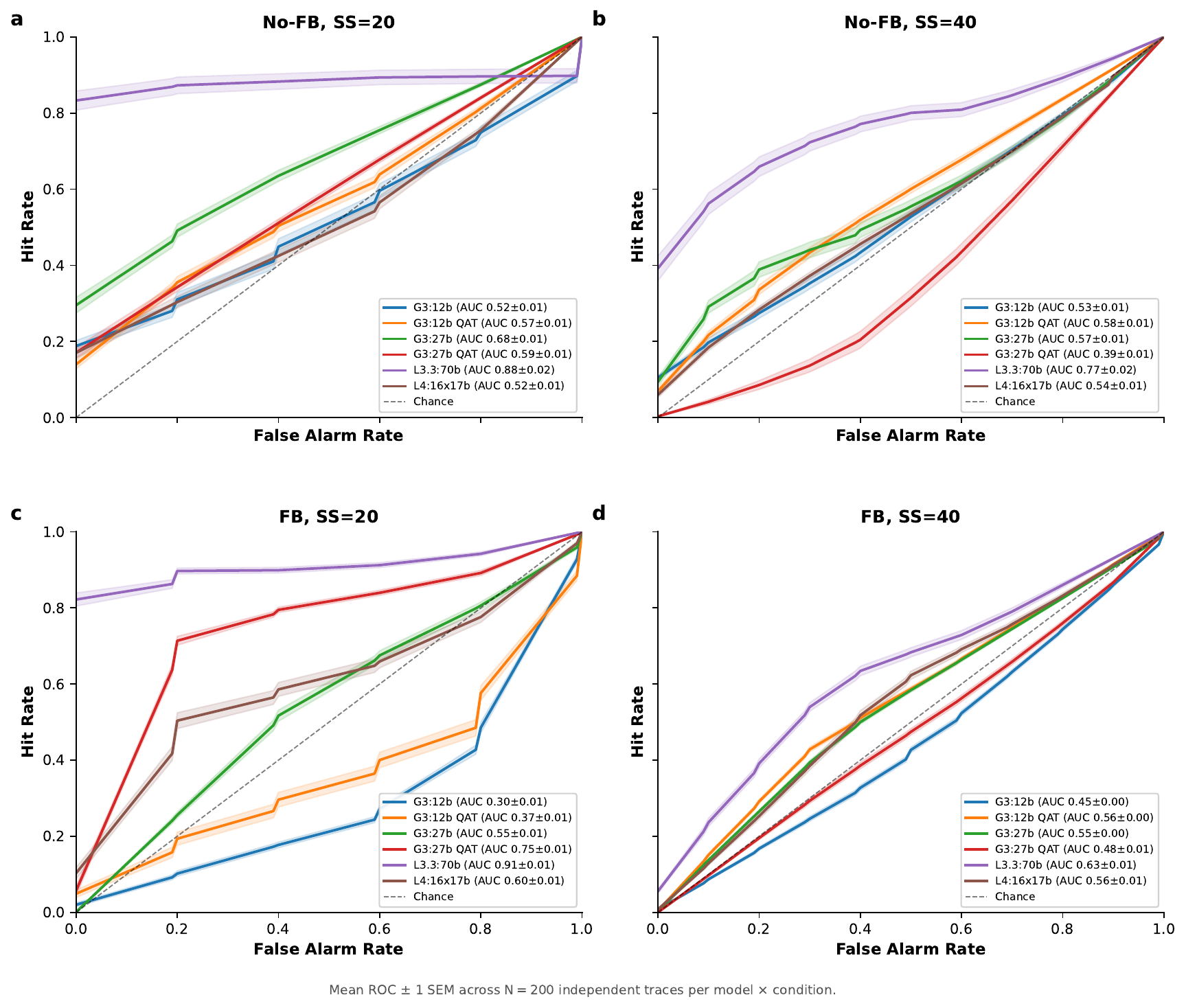}
  \caption{%
    \textbf{Without feedback, most models show near-chance source
    discrimination; corrective feedback reveals divergent
    discrimination profiles by model scale.}
    ROC curves for each model across the four Episodic-Chain
    conditions (panels: top-left~= set size~20, no feedback;
    top-right~= set size~20, feedback; bottom-left~= set size~40,
    no feedback; bottom-right~= set size~40, feedback).  Each curve
    shows one model; dashed diagonal~= chance (AUC~= 0.50).  Curves
    computed from signed confidence ratings.
    AUC per model reported in the legend.
    Without feedback, most models cluster near the diagonal,
    confirming near-chance source discrimination under genuine
    episodic delay.  This coexists with above-chance accuracy in
    Figure~4 because a model can achieve above-chance accuracy by
    being biased toward the more frequent source type without
    expressing confidence that tracks trial-level correctness.
    Llama3.3:70b is the exception, achieving well above-chance AUC
    in both no-feedback conditions, consistent with its elevated
    $d'$ in Figure~6.
    Under feedback, models diverge sharply.  Llama3.3:70b achieves
    its highest AUC at set size~20 under feedback and declines at
    the larger set size.  Gemma3:12b falls below chance at
    set size~20 under feedback, producing an inverted ROC curve that
    reflects systematic response reversal: the model's confidence
    ratings predict incorrect responses better than correct ones,
    confirming the negative $d'$ observed in Figure~6 and indicating
    that feedback disrupted its source--response mapping rather than
    improving it.  At set size~40, the feedback benefit is attenuated
    for most models.
  }
  \label{fig:s9}
\end{figure}

\clearpage


\section*{Experiment 2: CLM for Confidence Ratings}

Confidence ratings (ordinal, 1--6) from the Episodic-Chain paradigm were analyzed
using a Cumulative Link Model \cite{christensen2023ordinal} (CLM; logit link,
flexible thresholds) via R's \texttt{ordinal} package.  The interactive model
($\texttt{confidence} \sim \texttt{source} \times \texttt{setsize} +
 \texttt{source} \times \texttt{fb\_exp} + \texttt{setsize} \times \texttt{fb\_exp} +
 \texttt{model} + \texttt{accuracy} + \texttt{reading\_hallucination} +
 \texttt{rating\_cen} + \texttt{order\_c}$)
was compared to additive and null specifications by likelihood-ratio tests.
Full R output is archived in the project repository (see Code and Data Availability).

\begin{table}[htbp]
\caption{Experiment 2 Confidence Ratings: CLM Fixed-Effect Estimates (Interactive Model)}
\label{tab:s1_clm}
\small
\begin{tabular}{lrrrrrr}
\toprule
Term & $b$ & $SE$ & $z$ & $p$ & OR & 95\% CI \\
\midrule
source: imagined       &  0.098 & 0.033 &  2.938 &    .003 & 1.103 & [1.033, 1.178] \\
setsize: 40            & -0.057 & 0.029 & -1.982 &    .048 & 0.945 & [0.893, 0.999] \\
feedback: True         & -0.149 & 0.032 & -4.701 & $<$.001 & 0.862 & [0.810, 0.917] \\
model: Gemma3:12b-QAT  &  0.546 & 0.027 & 19.922 & $<$.001 & 1.726 & [1.636, 1.822] \\
model: Gemma3:27b      &  2.219 & 0.032 & 69.923 & $<$.001 & 9.198 & [8.645, 9.790] \\
model: Gemma3:27b-QAT  &  2.882 & 0.037 & 78.052 & $<$.001 & 17.846 & [16.606, 19.192] \\
model: Llama3.3:70b    &  0.922 & 0.028 & 33.017 & $<$.001 & 2.515 & [2.381, 2.656] \\
model: Llama4:16x17b   & -0.815 & 0.025 & -33.126 & $<$.001 & 0.442 & [0.422, 0.464] \\
accuracy: correct      &  0.030 & 0.016 &  1.858 &    .063 & 1.031 & [0.998, 1.065] \\
reading hallucination  &  0.000 & 0.025 &  0.019 &    .985 & 1.000 & [0.952, 1.051] \\
relatedness (centered) &  0.011 & 0.000 & 25.262 & $<$.001 & 1.011 & [1.010, 1.012] \\
order: imagined-first  &  0.132 & 0.016 &  8.429 & $<$.001 & 1.141 & [1.107, 1.177] \\
\midrule
source $\times$ setsize40      &  0.142 & 0.033 &  4.276 & $<$.001 & 1.153 & [1.080, 1.230] \\
source $\times$ feedback       & -0.503 & 0.032 & -15.791 & $<$.001 & 0.605 & [0.568, 0.644] \\
setsize40 $\times$ feedback    & -0.583 & 0.033 & -17.510 & $<$.001 & 0.558 & [0.523, 0.596] \\
\bottomrule
\end{tabular}

\vspace{6pt}
\noindent\textit{Note.} CLM = Cumulative Link Model (logit link; $N = 72{,}000$;
Nagelkerke $R^2 = .33$). OR = odds ratio; 95\% CI = profile likelihood interval.
Intercepts: LRT interactive vs.\ additive $\chi^2(3) = 572.73$,
$p < .001$; additive vs.\ null $\chi^2(12) = 24{,}645.09$, $p < .001$;
interactive vs.\ null $\chi^2(15) = 25{,}217.82$, $p < .001$.
Reference levels: source = perceived; setsize = 20; feedback = absent;
model = Gemma3:12b; accuracy = incorrect; reading hallucination = absent;
order = prompt option order (dummy-coded: 0 = external/perceived option listed
first in the source-attribution prompt [reference]; 1 = internal/imagined option
listed first).
\end{table}

\begin{table}[htbp]
\caption{Experiment 2 Confidence Ratings: Bonferroni-Corrected Post-Hoc Contrasts (CLM)}
\label{tab:s_clm_posthoc}
\small
\begin{tabular}{lrrrr}
\toprule
Contrast & Estimate & $SE$ & $z$ & $p$ \\
\midrule
\multicolumn{5}{l}{\textit{Source $\times$ Feedback}} \\
Perceived/No FB $-$ Imagined/No FB   & $-0.072$ & 0.011 &  $-6.49$ & $<$.001 \\
Perceived/No FB $-$ Perceived/FB     &  0.225   & 0.011 &   19.93  & $<$.001 \\
Perceived/No FB $-$ Imagined/FB      &  0.411   & 0.014 &   29.13  & $<$.001 \\
Imagined/No FB $-$ Perceived/FB      &  0.297   & 0.012 &   24.57  & $<$.001 \\
Imagined/No FB $-$ Imagined/FB       &  0.483   & 0.012 &   40.49  & $<$.001 \\
Perceived/FB $-$ Imagined/FB         &  0.185   & 0.015 &   12.73  & $<$.001 \\
\midrule
\multicolumn{5}{l}{\textit{Source $\times$ Set Size}} \\
Perceived/20 $-$ Imagined/20         &  0.082   & 0.014 &    5.97  & $<$.001 \\
Perceived/20 $-$ Perceived/40        &  0.183   & 0.011 &   16.20  & $<$.001 \\
Perceived/20 $-$ Imagined/40         &  0.214   & 0.013 &   16.67  & $<$.001 \\
Imagined/20 $-$ Perceived/40         &  0.101   & 0.013 &    7.66  & $<$.001 \\
Imagined/20 $-$ Imagined/40          &  0.132   & 0.012 &   11.23  & $<$.001 \\
Perceived/40 $-$ Imagined/40         &  0.031   & 0.012 &    2.56  &    .063 \\
\bottomrule
\end{tabular}

\vspace{6pt}
\noindent\textit{Note.} Estimated marginal mean contrasts from the
interactive CLM (\texttt{emmeans}), on the latent (logit) scale,
averaged over the levels of the remaining predictors; $p$ values
Bonferroni-corrected for six tests within each family.  FB = feedback
present; No FB = feedback absent; 20/40 = set size.
\end{table}


\clearpage
\section*{Experiment 2: Recognition Accuracy GLMM}

Source-attribution accuracy on test-phase trials ($N = 72{,}000$) was
modeled with a binomial GLMM (logit link) with a random intercept for
trace (\texttt{accuracy $\sim$ source\_test * setsize + source\_test *
fb\_exp + setsize * fb\_exp + model + order\_c + rating\_cen +
(1 | trace)}).  Fixed-effect estimates for the interactive model are
given in Table~\ref{tab:s_acc_glmm}; omnibus Type-II tests and the
back-transformed source contrast are reported in the main text.

\begin{table}[htbp]
\caption{Experiment 2 Recognition Accuracy: GLMM Fixed-Effect Estimates (Interactive Model)}
\label{tab:s_acc_glmm}
\small
\begin{tabular}{lrrrrr}
\toprule
Term & $b$ & $SE$ & $z$ & $p$ & OR [95\% CI] \\
\midrule
source: imagined           & $-1.766$ & 0.035 & $-50.944$ & $<$.001 & 0.171 [0.160, 0.183] \\
setsize: 40                & $-0.042$ & 0.034 &  $-1.222$ &    .222 & 0.959 [0.897, 1.026] \\
feedback: present          & $-0.908$ & 0.036 & $-25.036$ & $<$.001 & 0.403 [0.376, 0.433] \\
model: Gemma3:12b-QAT      &  0.202   & 0.032 &   6.264   & $<$.001 & 1.224 [1.149, 1.304] \\
model: Gemma3:27b          &  0.474   & 0.032 &  14.668   & $<$.001 & 1.606 [1.508, 1.712] \\
model: Gemma3:27b-QAT      &  0.391   & 0.032 &  12.101   & $<$.001 & 1.479 [1.388, 1.576] \\
model: Llama3.3:70b        &  1.510   & 0.036 &  42.517   & $<$.001 & 4.528 [4.224, 4.855] \\
model: Llama4:16x17b       &  0.377   & 0.032 &  11.752   & $<$.001 & 1.458 [1.370, 1.553] \\
order: imagined-first      &  0.014   & 0.019 &   0.738   &    .461 & 1.014 [0.977, 1.052] \\
relatedness (centered)     &  0.006   & 0.000 &  13.099   & $<$.001 & 1.006 [1.005, 1.007] \\
\midrule
source $\times$ setsize40  & $-0.089$ & 0.035 &  $-2.586$ &    .010 & 0.915 [0.855, 0.979] \\
source $\times$ feedback   &  1.734   & 0.033 &  52.289   & $<$.001 & 5.663 [5.307, 6.043] \\
setsize40 $\times$ feedback & $-0.179$ & 0.039 & $-4.549$ & $<$.001 & 0.837 [0.775, 0.903] \\
\bottomrule
\end{tabular}

\vspace{6pt}
\noindent\textit{Note.} Binomial GLMM (logit link; $N = 72{,}000$ test-phase
trials; random intercept for trace).  Interactive vs.\ additive model:
$\Delta\chi^2(3) = 2868.2$, $p < .001$; $R^2_m = .150$, $R^2_c = .176$,
ICC $= .031$.  Reference levels: source = perceived; setsize = 20;
feedback = absent; model = Gemma3:12b.  Coefficient rows for source,
setsize, and feedback are simple effects at reference levels of the
interacting factors; the omnibus Type-II main-effect tests reported in
the main text (e.g., set size $\chi^2(1) = 79.8$, $p < .001$) average
over those levels and can therefore differ from the coefficient-level
tests shown here.  Relatedness (mean-centered) is per unit of the
0--100 rating scale.
\end{table}


\clearpage
\section*{Experiment 2: Reading Hallucination GLMM}

Reading hallucinations on perceived trials ($N = 36{,}000$) were modeled
with a binomial GLMM (logit link) with a random intercept for trace
(\texttt{read\_hallucination $\sim$ setsize * fb\_exp + model + order\_c
+ (1 | trace)}).  Fixed-effect estimates for the interactive model are
given in Table~\ref{tab:s_rh_glmm}; the set size and feedback main
effects and their interaction are reported in the main text.

\begin{table}[htbp]
\caption{Experiment 2 Reading Hallucinations: GLMM Fixed-Effect Estimates (Interactive Model)}
\label{tab:s_rh_glmm}
\small
\begin{tabular}{lrrrrr}
\toprule
Term & $b$ & $SE$ & $z$ & $p$ & OR [95\% CI] \\
\midrule
Set size (40)              &  0.44   & 0.07 &   6.18  & $<$.001 & 1.56 [1.35, 1.79] \\
Feedback (present)         &  1.31   & 0.07 &  17.81  & $<$.001 & 3.72 [3.22, 4.30] \\
model: Gemma3:12b-QAT      &  0.97   & 0.07 &  13.07  & $<$.001 & 2.64 [2.28, 3.06] \\
model: Gemma3:27b          & $-0.89$ & 0.08 & $-11.68$ & $<$.001 & 0.41 [0.35, 0.48] \\
model: Gemma3:27b-QAT      & $-0.73$ & 0.08 &  $-9.67$ & $<$.001 & 0.48 [0.42, 0.56] \\
model: Llama3.3:70b        & $-6.25$ & 0.19 & $-33.27$ & $<$.001 & 0.002 [0.001, 0.003] \\
model: Llama4:16x17b       & $-1.90$ & 0.08 & $-23.92$ & $<$.001 & 0.15 [0.13, 0.17] \\
Order                      &  0.05   & 0.05 &   1.10  &    .270 & 1.05 [0.96, 1.16] \\
Set size $\times$ feedback &  0.43   & 0.10 &   4.45  & $<$.001 & 1.54 [1.28, 1.87] \\
\bottomrule
\end{tabular}

\vspace{6pt}
\noindent\textit{Note.} Binomial GLMM (logit link; $N = 36{,}000$ perceived
trials; random intercept for trace).  Reference levels: set size = 20;
feedback = absent; model = Gemma3:12b.  Set size $\times$ feedback
Type-II Wald $\chi^2(1) = 19.8$, $p < .001$.
In contrast with Experiment~1, where all four larger models hallucinated significantly less than the Gemma3:12b baseline and the 12B quantization-aware variant did not differ from it, under
episodic load the 12B quantization-aware variant hallucinated at
significantly \textit{higher} odds than its non-QAT counterpart
(OR $= 2.64$), and the 27B quantization-aware variant at numerically
higher odds than Gemma3:27b (0.48 vs.\ 0.41): quantization-aware
training conferred no consistent protection against reading
hallucinations across memory paradigms.
\end{table}


\clearpage
\section*{Experiment 1: Relatedness Rating Analysis}

Mean-centered relatedness ratings (rating\_cen; $N = 3{,}456$) were analyzed using
OLS regression with heteroskedasticity-consistent (HC3) standard errors
($\texttt{rating\_cen} \sim \texttt{C(accuracy)} + \texttt{C(source)} +
\texttt{C(memory)} + \texttt{C(model)} + \texttt{C(read\_hallucination)} +
\texttt{C(order)}$), yielding $R^2 = .132$, adjusted $R^2 = .129$,
$F(10, 3445) = 47.572$, $p < .001$.  To partition variance among predictors,
the model was refit without robust standard errors and a Type-II ANOVA was
computed; partial $\eta^2$ values are reported in Table~\ref{tab:s2_rr_anova}.
Source type was the dominant predictor (imagined word pairs rated as more
related than perceived pairs, by 12.8 points on average), followed by
model identity and memory
architecture; accuracy, reading hallucination, and response-option order
were not reliable predictors of relatedness ratings.  Figure~\ref{fig:s2}
shows the corresponding descriptive means.

\begin{table}[htbp]
\caption{Experiment 1 Relatedness Ratings: Type-II ANOVA and Partial $\eta^2$}
\label{tab:s2_rr_anova}
\small
\begin{tabular}{lrrrrr}
\toprule
Term & $SS$ & $df$ & $F$ & $p$ & Partial $\eta^2$ \\
\midrule
Accuracy             &    174.33 &    1 &   0.413 &    .521 & .0001 \\
Source               & 122{,}649.70 &    1 & 290.488 & $<$.001 & .0778 \\
Memory               &  25{,}075.60 &    1 &  59.390 & $<$.001 & .0169 \\
Model                &  57{,}704.75 &    5 &  27.334 & $<$.001 & .0382 \\
Reading hallucination &    26.33 &    1 &   0.062 &    .803 & .0000 \\
Order                &    630.85 &    1 &   1.494 &    .222 & .0004 \\
Residual             & 1{,}454{,}548.00 & 3{,}445 &     --- &    --- &  --- \\
\bottomrule
\end{tabular}

\vspace{6pt}
\noindent\textit{Note.} Type-II sums of squares from \texttt{statsmodels.stats.anova.anova\_lm}
on the OLS fit (without HC3 correction); partial $\eta^2 = SS_{\text{effect}} /
(SS_{\text{effect}} + SS_{\text{Residual}})$.  Source = perceived vs.\ imagined;
Memory = Single-Turn vs.\ Trial-Chain; Model = six-level model factor.
\end{table}


\clearpage
\section*{Experiment 1: Random-Effects Sensitivity Analysis}

To confirm that the primary Experiment~1 inferences reported in the main text do not
depend on treating \texttt{model} (LLM identity) as a fixed categorical predictor, the
three primary Experiment~1 models (reading hallucination, recognition accuracy, and relatedness rating) were re-estimated with \texttt{model} instead entered as a random
intercept, \texttt{(1 | model)}, and all other predictors unchanged.  This is a more
conservative specification: it treats the six tested LLMs as a sample from a broader
population of architectures rather than as the fixed set of conditions studied, at the
cost of substantial imprecision in the between-model variance estimate given only six
model levels.

\subsubsection*{Reading Hallucination (perceived-source trials, $N = 1{,}728$)}

\begin{table}[htbp]
\caption{Experiment 1 Random-Effects Sensitivity: Reading Hallucination GLMM}
\label{tab:s3_re_rh}
\small
\begin{tabular}{lrrrrr}
\toprule
Term & $b$ & $SE$ & $z$ & $p$ & $OR$ [95\% CI] \\
\midrule
Memory (Trial-Chain)   & $-1.80$  & 0.16 & $-10.92$ & $<$.001 & 0.17 [0.12, 0.23] \\
Order                  & 0.33     & 0.14 & 2.31     & .021    & 1.40 [1.05, 1.85] \\
Relatedness (centered) & $-0.006$ & 0.003& $-2.07$  & .039    & 0.99 [0.99, 1.00] \\
\bottomrule
\end{tabular}
\vspace{4pt}
\noindent\textit{Note.} \texttt{read\_hallucination $\sim$ memory + order\_c + rating\_cen +
(1 | model)}; binomial GLMM.  ICC$_{\text{model}} = .361$ ($\sigma^2_{\text{model}} = 1.86$,
$\sigma^2_{\text{total}} = 5.15$, latent scale).  Type-II Wald tests: memory
$\chi^2(1) = 119.3$; order $\chi^2(1) = 5.35$; relatedness $\chi^2(1) = 4.28$
(all $p < .05$).
\end{table}

\subsubsection*{Recognition Accuracy (perceived-source trials, $N = 1{,}728$)}

\begin{table}[htbp]
\caption{Experiment 1 Random-Effects Sensitivity: Recognition Accuracy GLMM}
\label{tab:s3_re_acc}
\small
\begin{tabular}{lrrrrr}
\toprule
Term & $b$ & $SE$ & $z$ & $p$ & $OR$ [95\% CI] \\
\midrule
Memory (Trial-Chain)   & $-1.23$  & 0.23 & $-5.26$  & $<$.001 & 0.29 [0.19, 0.46] \\
Reading hallucination  & $-3.86$  & 0.24 & $-15.80$ & $<$.001 & 0.02 [0.01, 0.03] \\
Order                  & $-0.56$  & 0.18 & $-3.13$  & .002    & 0.57 [0.40, 0.81] \\
Relatedness (centered) & $-0.003$ & 0.004& $-0.90$  & .370    & 1.00 [0.99, 1.00] \\
\bottomrule
\end{tabular}
\vspace{4pt}
\noindent\textit{Note.} \texttt{accuracy $\sim$ memory + read\_hallucination + order\_c +
rating\_cen + (1 | model)}; binomial GLMM.  ICC$_{\text{model}} = .491$
($\sigma^2_{\text{model}} = 3.18$, $\sigma^2_{\text{total}} = 6.47$, latent scale).
Type-II Wald tests: memory $\chi^2(1) = 27.68$; reading hallucination
$\chi^2(1) = 249.7$; order $\chi^2(1) = 9.77$ (all $p < .01$); relatedness
$\chi^2(1) = 0.81$, $p = .370$ (n.s.).
\end{table}

\subsubsection*{Relatedness Rating (all trials, $N = 3{,}456$)}

\begin{table}[htbp]
\caption{Experiment 1 Random-Effects Sensitivity: Relatedness Rating LMM}
\label{tab:s3_re_rr}
\small
\begin{tabular}{lrrrr}
\toprule
Term & $b$ & $SE$ & $t$ & $p$ \\
\midrule
Source (imagined)      & 12.76   & 0.75 & 17.05   & $<$.001 \\
Accuracy (correct)     & $-1.05$ & 1.69 & $-0.62$ & .533 \\
Memory (Trial-Chain)   & $-5.50$ & 0.71 & $-7.70$ & $<$.001 \\
Reading hallucination  & 0.48    & 1.69 & 0.28    & .779 \\
Order                  & 0.86    & 0.70 & 1.22    & .222 \\
\bottomrule
\end{tabular}
\vspace{4pt}
\noindent\textit{Note.} \texttt{rating\_cen $\sim$ source + accuracy + memory +
read\_hallucination + order\_c + (1 | model)}; Gaussian LMM.  ICC$_{\text{model}} = .044$
($\sigma^2_{\text{model}} = 19.42$, $\sigma^2_{\text{total}} = 441.64$).  Type-II Wald
tests: source $\chi^2(1) = 290.8$, $p < .001$; memory $\chi^2(1) = 59.3$, $p < .001$;
accuracy, reading hallucination, and order all $p > .2$ (n.s.).
\end{table}

\subsubsection*{Variance Attributable to Model Identity}

\begin{table}[htbp]
\caption{Experiment 1 Random-Effects Sensitivity: Between-Model Variance Components}
\label{tab:s3_re_icc}
\small
\begin{tabular}{lrrr}
\toprule
Outcome & $\sigma^2_{\text{model}}$ & $\sigma^2_{\text{total}}$ & ICC$_{\text{model}}$ \\
\midrule
Reading hallucination & 1.86  & 5.15   & .361 \\
Recognition accuracy  & 3.18  & 6.47   & .491 \\
Relatedness rating    & 19.42 & 441.64 & .044 \\
\bottomrule
\end{tabular}
\vspace{4pt}
\noindent\textit{Note.} ICC$_{\text{model}} = \sigma^2_{\text{model}} / \sigma^2_{\text{total}}$,
the proportion of total outcome variance attributable to LLM-architecture identity.  With
only six model levels, these variance estimates carry substantial uncertainty and should
be treated as indicative rather than precise.
\end{table}

Across all three models, the primary fixed-effect inferences reported in the main text
replicated in direction and significance under the random-effects specification: the
Trial-Chain reduction in reading hallucinations ($z = -10.92$, $p < .001$), the detrimental
effect of reading hallucinations on recognition accuracy ($z = -15.80$, $p < .001$), and
the source advantage in relatedness ratings ($t = 17.05$, $p < .001$).  The between-model
variance was low for relatedness rating (ICC $= .044$) but moderate-to-high for reading
hallucination and recognition accuracy (ICC $= .361$ and $.491$, respectively), indicating
that LLM-architecture identity accounts for a non-trivial share of variance in these two
outcomes even under the more conservative random-effects specification.  This does not
change the direction or significance of the effects reported above, but it means
fixed-effects inferences about memory format and reading-hallucination effects should be
interpreted as specific to the six tested architectures rather than as estimates
generalizing to the broader population of LLM designs; with only six model levels, more
precise population-level estimates would require a substantially larger sample of model
architectures.


\clearpage
\section*{Prompt Architecture}

\noindent
Complete raw simulation records (one \texttt{.psyscan} JSON file per
trace, capturing the system message, all per-trial human turns, and
the model's responses) are archived in the project repository
(blinded for review; see the Code and Data Availability statement in
the main text for the camera-ready repository URL).
The root-relative paths within the repository are:

{\setlength{\itemsep}{0pt}\setlength{\parsep}{0pt}\setlength{\topsep}{2pt}
\begin{itemize}
  \item Experiment~1, Single-Turn: \texttt{data/external/rm\_2op/}
  \item Experiment~1, Trial-Chain: \texttt{data/external/rm\_2op\_tc/}
  \item Experiment~2, no-feedback ($n=20$): \texttt{data/external/exp2/5t\_100/}
  \item Experiment~2, no-feedback ($n=40$): \texttt{data/external/exp2/10t\_100/}
  \item Experiment~2, feedback ($n=20$): \texttt{data/external/exp2/5t\_fb\_100/}
  \item Experiment~2, feedback ($n=40$): \texttt{data/external/exp2/10t\_fb\_100/}
\end{itemize}}

All prompts were serialized as JSON and delivered via the PsychScanner API
(v0.1) using LangChain's chat-message format (\textit{role}: \textit{system}
or \textit{human}).
In the templates below, concrete stimuli are replaced by angle-bracketed
placeholders (e.g., \texttt{\textcolor{placeholdercolor}{\textbf{<word\_1>}}});
commented lines (\texttt{//~\textit{...}}) are annotations added here and were
not present in the actual prompts.

\bigskip
\noindent\textbf{Task components referenced across all prompts:}

\begin{list}{}{%
  \setlength{\leftmargin}{1em}%
  \setlength{\labelwidth}{0pt}%
  \setlength{\labelsep}{0.5em}%
  \setlength{\itemindent}{0pt}%
  \setlength{\itemsep}{0pt}%
  \setlength{\parsep}{0pt}%
  \setlength{\topsep}{2pt}%
  \let\makelabel\descriptionlabel
}
  \item[\textbf{word-pair}] Generate or reproduce \textit{word\_2}: if \textit{word\_2} is blank, imagine a novel
        single English word (not word\_1, not used in any prior trial, not a compound,
        symbol, or number); if \textit{word\_2} is provided, report it verbatim.
        (Telegraphic style is intentional; prompts were designed to minimize token
        length while remaining unambiguous to instruction-tuned models.)
  \item[\textbf{relatedness}] Rate semantic relatedness of \textit{word\_1} and \textit{word\_2} on
        a continuous 0--100 scale (0 = unrelated, 100 = highly related;
        phonetic, semantic, or categorical similarity).
  \item[\textbf{judgment}] Attribute the origin of \textit{word\_2}: \texttt{"internal"} if imagined;
        \texttt{"external"} if provided.
  \item[\textbf{confidence}] Rate certainty in the source judgment on a 1--6 scale
        (1 = not at all confident, 6 = highly confident).
\end{list}

\subsection*{S.PA\quad Experiment 1 --- Single-Turn}

The model received all four task components (\textit{word-pair},
\textit{relatedness}, \textit{judgment}, \textit{confidence})
simultaneously in a single system message, and was required to produce
all four fields in one JSON response per trial.  Each trial began a
fresh context window.

\medskip
\noindent\textbf{System message (role: system)} ---
\textit{abbreviated; full text in \texttt{rm\_2op/} files}

\begin{lstlisting}[style=promptjson]
{
  "TASK CONTEXT": {
    "definition": [
      "You are a helpful participant performing a task with four
       different components for successful response.",
      "Four components: [WORD_PAIR_TASK, RELATEDNESS_RATING,
       JUDGE_GENERATION_TYPE, CONFIDENCE_ON_JUDGEMENT_OF_GENERATION_TYPE]",
      "Follow all instructions for each component."
    ],
    "INSTRUCTIONS": {
      "WORD_PAIR_TASK":  { ... },   // see component description above
      "RELATEDNESS_RATING_BETWEEN_WORD_1_AND_WORD_2": { ... },
      "JUDGE_GENERATION_TYPE_OF_WORD_2": {
        "options": { "enum": ["external", "internal"] },
        "option_descriptions": {
          "external": "word_2 was provided in the task as an English word.",
          "internal": "word_2 was imagined by you to complete the word-pair."
        }
      },
      "CONFIDENCE_ON_JUDGEMENT_OF_GENERATION_TYPE": {
        "enum": [1, 2, 3, 4, 5, 6],
        "confidence_scale": {
          "1": "Not at all confident.",  "2": "Slightly confident.",
          "3": "Moderately confident.", "4": "Fairly confident.",
          "5": "Very confident.",       "6": "Highly confident."
        }
      }
    },
    "response_format": {
      "Word_2":      "<generated or reproduced word>",
      "Rating":      "<relatedness rating, 0-100>",
      "Judgement":   "<internal | external>",
      "Confidence":  "<confidence level, 1-6>"
    }
  }
}
\end{lstlisting}

\medskip
\noindent\textbf{Human turn --- imagined trial (word\_2 blank):}

\begin{lstlisting}[style=promptjson]
{"Word_Pair": {"word_1": "<word_1>", "word_2": "________"}}
\end{lstlisting}

\noindent\textbf{Human turn --- perceived trial (word\_2 provided):}

\begin{lstlisting}[style=promptjson]
{"Word_Pair": {"word_1": "<word_1>", "word_2": "<word_2>"}}
\end{lstlisting}

\noindent\textbf{Expected model response (single JSON object, one turn):}

\begin{lstlisting}[style=promptjson]
{"Word_2": "<word>", "Rating": <0-100>,
 "Judgement": "<internal|external>", "Confidence": <1-6>}
\end{lstlisting}

\subsection*{S.PB\quad Experiment 1 --- Trial-Chain}

In the Trial-Chain condition the same four components were delivered
\textit{sequentially}: each word pair occupied four consecutive turns in a
single growing conversation, one component instruction + query per turn.
The system message contained only the abbreviated role definition; full
component instructions were provided inline with each turn's query.

\medskip
\noindent\textbf{System message (role: system)} ---
\textit{abbreviated; full text in \texttt{rm\_2op\_tc/} files}

\begin{lstlisting}[style=promptjson]
{
  "TASK CONTEXT": {
    "definition": [
      "You are a helpful participant performing a task with four
       different components for successful response.",
      "Four components: [WORD_PAIR_TASK, RELATEDNESS_RATING,
       JUDGE_GENERATION_TYPE, CONFIDENCE_ON_JUDGEMENT_OF_GENERATION_TYPE]"
    ]
  }
}
\end{lstlisting}

\medskip
\noindent\textbf{Turn 1 of 4 --- word completion:}

\begin{lstlisting}[style=promptjson]
{
  "instructions": { "WORD_PAIR_TASK": { ... } },  // inline component instructions
  "Word_Pair":    { "word_1": "<word_1>", "word_2": "________" },
  "Query": "Please report the Word_2 as per instructions. Word 2 = "
}
// Expected model response: {"response": {"Word_2": "<generated word>"}}
\end{lstlisting}

\noindent\textbf{Turn 2 of 4 --- relatedness rating:}

\begin{lstlisting}[style=promptjson]
{
  "instructions": { "RELATEDNESS_RATING_BETWEEN_WORD_1_AND_WORD_2": { ... } },
  "Query": "Please report Relatedness Rating between the two words in
            the earlier word-pair as per instructions. Relatedness Rating = "
}
// Expected model response: {"response": {"Relatedness_Rating": <0-100>}}
\end{lstlisting}

\noindent\textbf{Turn 3 of 4 --- source judgment:}

\begin{lstlisting}[style=promptjson]
{
  "instructions": { "JUDGE_GENERATION_TYPE_OF_WORD_2": { ... } },
  "Query": "Please select and report the Judgment of word_2 value
            regarding generation type using the earlier word-pair.
            Select one of the generation type options as per the
            instructions. Judgment = "
}
// Expected model response: {"response": {"Judgment": "<internal|external>"}}
\end{lstlisting}

\noindent\textbf{Turn 4 of 4 --- confidence rating:}

\begin{lstlisting}[style=promptjson]
{
  "instructions": { "CONFIDENCE_ON_JUDGEMENT_OF_GENERATION_TYPE": { ... } },
  "Query": "Please report the Confidence on your judgment being correct
            as per the confidence rating scale in your generation type
            judgment. Confidence = "
}
// Expected model response: {"response": {"Confidence": <1-6>}}
\end{lstlisting}

\subsection*{S.PC\quad Experiment 2 --- Episodic-Chain Encoding Phase}

In the Episodic-Chain paradigm (Experiment~2) the encoding phase
included only the two production components (\textit{word-pair} and
\textit{relatedness}).  Source judgment and confidence were withheld
until the test phase (see S.PD below).  All encoding trials for one
trace were presented as consecutive human turns in a single growing
conversation; the model's responses accumulated in the same context
window, which carried into the test phase.

\medskip
\noindent\textbf{Encoding system message (role: system)} ---
\textit{abbreviated; full text in \texttt{rm\_2op\_convo/} files}

\begin{lstlisting}[style=promptjson]
{
  "task_definition": [
    "You are a helpful participant performing a task with two
     different components for successful response.",
    "Two components: [WORD_PAIR_TASK,
     RELATEDNESS_RATING_BETWEEN_WORD_1_AND_WORD_2]",
    "Follow all instructions for each component."
  ],
  "INSTRUCTIONS": {
    "WORD_PAIR_TASK": { ... },              // as described in S.PA
    "RELATEDNESS_RATING_BETWEEN_WORD_1_AND_WORD_2": { ... }
  }
}
\end{lstlisting}

\noindent\textbf{Encoding stimulus --- imagined trial:}

\begin{lstlisting}[style=promptjson]
{"Word_Pair": {"word_1": "<word_1>", "word_2": "________"}}
\end{lstlisting}

\noindent\textbf{Encoding stimulus --- perceived trial:}

\begin{lstlisting}[style=promptjson]
{"Word_Pair": {"word_1": "<word_1>", "word_2": "<word_2>"}}
\end{lstlisting}

\noindent\textbf{Expected encoding response (two fields only):}

\begin{lstlisting}[style=promptjson]
{"Word_2": "<word>", "Rating": <0-100>}
\end{lstlisting}

\subsection*{S.PD\quad Experiment 2 --- Episodic-Chain Test Phase}

After all encoding trials were complete, the system message was updated
(in the same conversation) to introduce the test-phase task.
The test system message referenced the previously completed encoding,
and only the two attribution components (\textit{judgment} and
\textit{confidence}) were active.  Each test trial presented a single
\textit{Test\_Word} that had appeared as \textit{word\_1} in one of
the encoding pairs.

\medskip
\noindent\textbf{Test system message (role: system)} ---
\textit{abbreviated; full text in \texttt{rm\_2op\_convo/} files}

\begin{lstlisting}[style=promptjson]
{
  "task_definition": [
    "You are a helpful participant performing a task based on the
     previously completed word-pair task with two different
     components for successful response.",
    "You will be asked to make judgements about the word you
     reported as word_2 in the previous task based on the Test_Word.",
    "Two components: [JUDGE_GENERATION_TYPE_OF_WORD_2,
     CONFIDENCE_ON_JUDGEMENT_OF_GENERATION_TYPE]"
  ],
  "INSTRUCTIONS": {
    "JUDGE_GENERATION_TYPE_OF_WORD_2": {
      "definition": [
        "During the word-pair task, Test_Word would have appeared as
         word_1 in one of the pairs for which you reported the same word
         as word_2 or imagined it.",
        "Judge the nature of generation of the value of word_2:"
      ],
      "options":            { "enum": ["external", "internal"] },
      "option_descriptions": {
        "external": "Only if word_2 was provided in the task as an English word.",
        "internal": "Only if word_2 was imagined by you to complete the word-pair and replace the blank."
      }
    },
    "CONFIDENCE_ON_JUDGEMENT_OF_GENERATION_TYPE": { ... }  // as in S.PA
  }
}
\end{lstlisting}

\noindent\textbf{Test stimulus (word\_1 from encoding, no word\_2):}

\begin{lstlisting}[style=promptjson]
{"Test_Word": "<word_1_from_encoding_trial>"}
\end{lstlisting}

\noindent\textbf{Expected test response (two fields only):}

\begin{lstlisting}[style=promptjson]
{"Judgment": "<internal|external>", "Confidence": <1-6>}
\end{lstlisting}

\subsection*{S.PE\quad Experiment 2 --- Feedback Condition}

In the feedback condition, each trial's stimulus (both during encoding
and test) was prefixed with a three-component JSON feedback object
reporting performance on the immediately preceding trial.
The three components (\textit{response feedback}, \textit{rating feedback}, and \textit{overall feedback}) mapped
directly onto the three scored aspects of the prior trial's response.
The combined feedback + stimulus object constituted the single human
turn for that trial; no separate feedback turn was added.
The encoding and test system messages (S.PC, S.PD) were also extended
with three additional sentences present only in the feedback condition,
informing the model that feedback on previous trials would be provided
and instructing it to use that feedback to improve task performance
and respond as accurately as possible.

\medskip
\noindent\textbf{Feedback prefix structure (encoding phase, imagined trial shown):}

\begin{lstlisting}[style=promptjson]
{
  "Feedback on previous response": {
    "response feedback": "<CORRECT|INCORRECT message about Word_2 novelty>",
    "rating feedback":   "<CORRECT|INCORRECT message about rating range>",
    "overall feedback":  "<CORRECT|INCORRECT Answer!>"
  },
  "current_trial": {
    "Word_Pair": {"word_1": "<word_1>", "word_2": "________"}
  }
}
// For perceived trials, "word_2" is filled with the provided word.
// For test trials, "Word_Pair" is replaced by "Test_Word": "<word_1_from_encoding>"
\end{lstlisting}

\medskip
\noindent\textbf{Feedback message variants:}

\smallskip
\textit{Encoding --- perceived trial, response correct:}
\begin{lstlisting}[style=promptjson]
"response feedback": "CORRECT: Last trial was a perceived trial type,
  your 'Word_2' value correctly followed the trial instruction for
  the given trial."
\end{lstlisting}

\textit{Encoding --- perceived trial, response incorrect:}
\begin{lstlisting}[style=promptjson]
"response feedback": "INCORRECT: Last trial was a perceived trial type,
  your 'Word_2' value is INCORRECT because the second word in the
  previous trial word-pair does not match your 'response'. It might
  be also be incorrect due to poor formatting of your ANSWER.
  Carefully follow all the trial instructions related to perceived
  trials."
\end{lstlisting}

\noindent Imagined-trial encoding feedback follows an analogous
correct/incorrect structure, with INCORRECT reasons specific to
imagined trials (a non-novel second word, or a second word not found
in an English dictionary); perceived-trial examples are shown above
for brevity.

\textit{Test --- source identification correct (shown for internal source; analogous for external):}
\begin{lstlisting}[style=promptjson]
"response feedback": "CORRECT: Correct generation for 'Word_2' response
  was internal. Your 'Word_2' response followed the task instructions."
// For external: "...response was external. ..."
\end{lstlisting}

\textit{Test --- source identification incorrect (shown for internal source; analogous for external):}
\begin{lstlisting}[style=promptjson]
"response feedback": "INCORRECT: Correct generation for 'Word_2'
  response was internal. It might be also be incorrect due to poor
  formatting of your ANSWER. Carefully follow all the trial
  instructions about the 'response' options to be accurate on the
  given trial."
// For external: "...response was external. ..."
\end{lstlisting}

\noindent The \textit{rating feedback} and \textit{overall feedback} fields were identical
across all variants; the only observed \textit{rating feedback} message was:

\begin{lstlisting}[style=promptjson]
"rating feedback":  "CORRECT: Your 'rating' value is in the instructed
                    rating scale range."
// Encoding scale: 0-100; Test scale: 1-6 (confidence)
\end{lstlisting}

\noindent The \textit{overall feedback} field reflected the conjunction of both components:
\begin{lstlisting}[style=promptjson]
// Both correct:   "CORRECT Answer! Well Answered, Good Job!"
// Either wrong:   "INCORRECT Answer! Follow all the trial instructions
//                 more accurately to give correct response and rating."
\end{lstlisting}


\section*{Declaration of Generative AI and AI-Assisted Technologies in the Writing Process}

The authors used Claude (Anthropic, version claude-sonnet-4-6) during the preparation of this
manuscript. The tool was used in the paper text for language editing, clarity improvements,
\LaTeX{} typesetting, and code development. The authors reviewed, edited, and validated all AI-assisted outputs and
made all core intellectual and design decisions.


\bibliography{sn-bibliography}

\end{document}